\documentclass[journal]{IEEEtran}
\usepackage{amsmath,amsfonts}
\usepackage{array}
\usepackage[caption=false,font=normalsize,labelfont=sf,textfont=sf]{subfig}
\usepackage{textcomp}
\usepackage{stfloats}
\usepackage{url}
\usepackage{verbatim}
\usepackage{graphicx}
\usepackage[ruled,linesnumbered,algo2e]{algorithm2e}
\usepackage{algorithmic}
\usepackage{algorithm}
\usepackage{pifont}
\usepackage{multirow}

\usepackage[colorlinks,linkcolor=blue, anchorcolor=blue,citecolor=green]{hyperref}
\def\BibTeX{{\rm B\kern-.05em{\sc i\kern-.025em b}\kern-.08em
    T\kern-.1667em\lower.7ex\hbox{E}\kern-.125emX}}
\begin{document}
\title{Ultra-slender Coaxial Antagonistic Tubular Robot for Ambidextrous Manipulation}
\author{Qingxiang Zhao$^1$, Runfeng Zhu$^2$, Xin Zhong$^3$, Baitao Lin$^4$, Xiandi Wang$^1$, Xilong Hou$^5$, Jian Hu$^5$, Kang Li$^{1,*}$
\thanks{Corresponding author: Kang Li.
	
	This work is partially supported by  1·3·5 Project for Disciplines of Excellence and  also supported by Interdisciplinary Crossing and Integration of Medicine and Engineering for Talent Training Fund, West China Hospital, Sichuan University ( ZYYC21004).
	
Qingxiang Zhao, Xiandi Wang and Kang Li are both with the West China Hospital of Medicine, Sichuan University, Chengdu, China. (email: qingxiang.zhao@wchscu.cn, wxdfrank@scu.edu.cn and likang@wchscu.cn); Runfeng Zhu is with the Shien-Ming Wu School of Intelligent Engineering, South China University of Technology, Guangzhou, China. (email: wirun-feng.zhu@mail.scut.edu.cn); Xin Zhong is with the School of Mechanical and Electrical Engineering, University of Electronic Science and Technology of China, Chengdu, China. (email: xin\_zhong@std.uestc.edu.cn); Baitao Lin is with the School of Mechanical Engineering, Sichuan University, Chengdu, China. (email: linbaitao@stu.scu.edu.cn); Xilong Hou and Jian Hu are with the Centre for Artificial Intelligence and Robotics (CAIR) Hong Kong Institute of Science and Innovation, Chinese Academy of Sciences, Hong Kong, China. (email:xilong.hou@cair-cas.org.hk and hujian@cair-cas.org.hk).
}}

\maketitle

\begin{abstract}
As soft continuum manipulators characterize terrific compliance and maneuverability in narrow unstructured space, low stiffness and limited dexterity are two obvious shortcomings in practical applications. To address the issues, a novel asymmetric coaxial antagonistic tubular robot (CATR) arm with high stiffness has been proposed, where two asymmetrically patterned metal tubes were fixed at the tip end with a shift angle of $ 180^\circ $ and axial actuation force at the other end deforms the tube. Delicately designed and optimized steerable section and fully compliant section enable the soft manipulator high dexterity and stiffness. The basic kinetostatics model of a single segment was established on the basis of geometric and statics, and constrained optimization algorithm promotes finding the actuation inputs for a given desired task configuration. In addition, we have specifically built the design theory for the slits patterned on the tube surface, taking both bending angle and stiffness into account. Experiments demonstrate that the proposed robot arm is dexterous and has greater stiffness compared with same-size continuum robots. Furthermore, experiments also showcase the potential in minimally invasive surgery.
\end{abstract}


\section{Introduction}
As an important branch of
 robotics, continuum robots have been applied in various scenarios, including Minimally Invasive Surgery (MIS) \cite{wang2023novel, dupont2022continuum}, aerospace industry \cite{russo2023continuum}, and manipulating irregular shaped objects \cite{zhang2022novel, jalali2024dynamic}, owing to their slender body, safe contact, and dexterous motion in confined spaces. In recent decades, advanced functional materials to form the robot bodies \cite{zhang2022integrated}, tuneable stiffness designs \cite{shen2023design}, novel actuation mechanisms \cite{wang2023novel}, and advanced modelling/control strategies \cite{tummers2023cosserat} have been proposed in the research community of continuum robotics. For dexterous manipulation in narrow confined task space, such as MIS scenario, micro-scale/meso-scale flexible robot manipulator arms are required, but the challenges are twofold: 1) slender robot body contradicts actuation unit and instrument channel space, and 2) slender body characterizes low stiffness and has limited manipulation force at the end effector.
\begin{figure}[!t]
	\centering
	\includegraphics[width=1\linewidth]{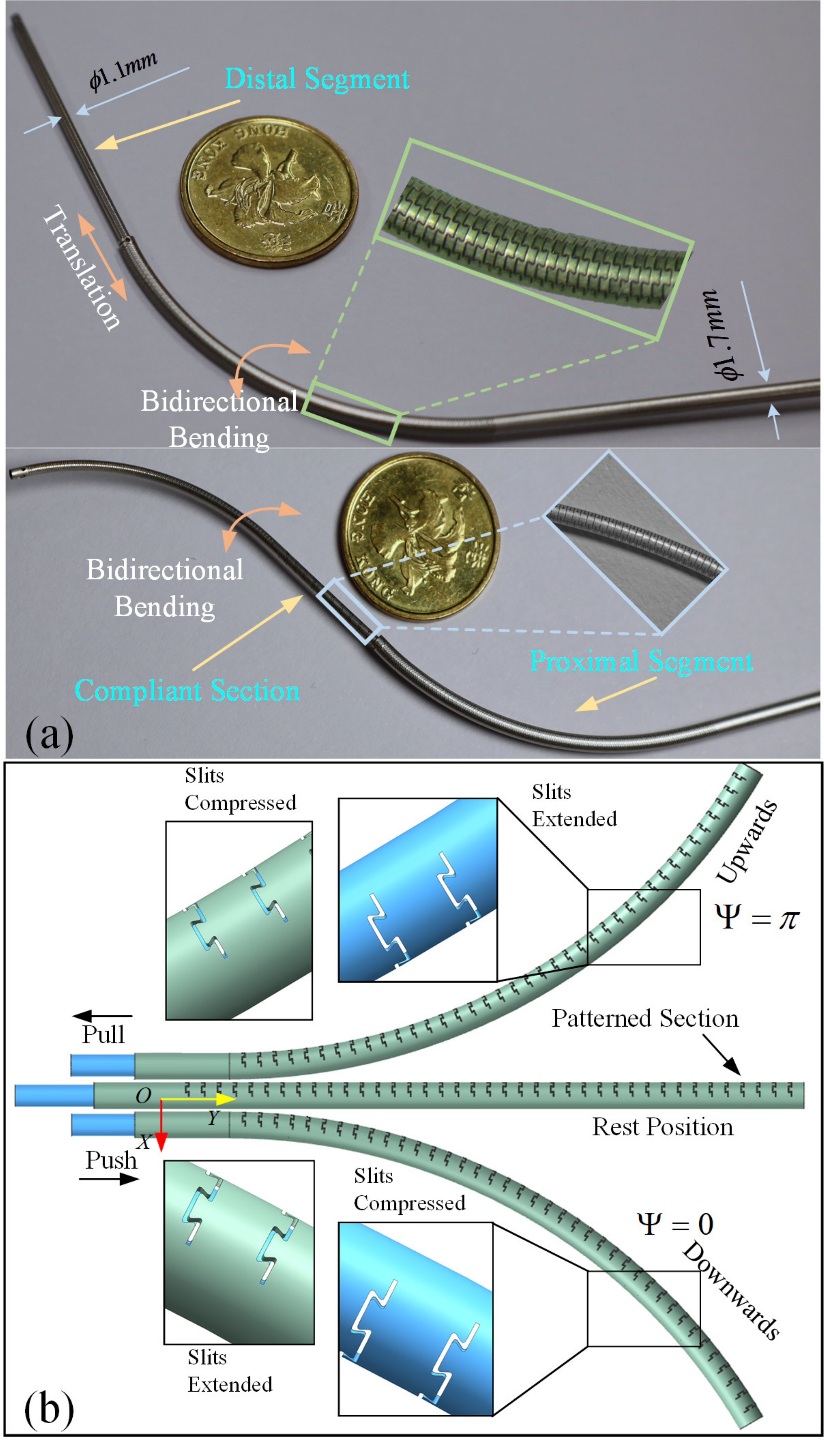}
	\caption{(a) Dual-segment CATR. (b) A single-segment CATR. It has two hollow thin-walled tubes, and each of them were patterned with dense tenon-mortise slits. They are fixed at the tip. Pushing/pulling the inner tube steers the patterned section to bend downwards/upwards. }
	\label{fig:basics}
\end{figure}

Pursuing novel actuation mechanisms and advanced fabrication methods of continuum robot body are always hot topics in recent decade. At present, conventional continuum robot consists of three secondary backbones for actuation, where elongation/contraction induced by tension (tendon-driven mechanism) or pressurized fluid (fluid-driven mechanism) deform the soft manipulator with two DoFs (direction angle and bending angle). For multi-segment designs, the actuation media for the distal segments, such as tendon sheath \cite{lilge2022kinetostatic} and pressurized fluid \cite{zhong2020recent}, should pass through the secondary backbones inside the proximal ones, which can not be slender ($<$2mm in diameter) and leaves little space for instrument channel. Additionally, some vertebrated designs employed central backbone to constrain the maximum length \cite{zhao2021reconstructing}. Extrinsic magnetic fields actuate magnets mounted on a continuum manipulator to realize overall deformation, but it suffers from environmental interference, low force, and bulky external magnetic field generator \cite{kim2019ferromagnetic}. Among them, tendon-driven continuum robots are most compact since the cable diameter could be 0.16mm \cite{kato2014tendon} while the sheath to reduce friction also needs space. Micro-scale  \cite{chitalia2020towards} steerable guidewires made of NiTi could reach an diameter of 0.78mm \cite{chitalia2018design}, but only two DoFs were achieved. Another category relies on intrinsic actuation. Compact dielectric actuators \cite{mc2020continuum}, pressurized fluid in antagonistic compliant chambers, tip follower actuation \cite{campisano2021closed} and growing soft robots \cite{tutcu2021quasi} are also not able to be miniaturized to comparable small dimension. Concentric tube robots are made of pre-curved NiTi tubes. Axial translation and rotation with respective to adjacent tubes deform the whole soft manipulator, and the small tube dimension provides them with potential in surgical application \cite{nwafor2023design}. However, the motion of upscale tubes is coupled with the inner ones, and the limited stiffness does not suited for with-load manipulation. This work proposed a novel design where the outer diameter of the dual-segment continuum robot is only 1.7mm and the tool channel is of 0.6mm in diameter. Each segment has 3 DoFs and the overall stiffness is much higher than conventional concentric tube robots, as shown in Fig. \ref{fig:basics} (a).

In addition to the ultra-slender body and larger tool channel space, sufficient stiffness to compensate external loads and acting force at the tip are both desired. Tuneable stiffness is desirable, as low stiffness ensures compliance and higher stiffness promotes effective manipulation. Owing to the essential redundancy, higher sum of actuation forces leads to larger stiffness because the secondary backbones are tensioned and the shape configuration depends on relative actuation force from the secondary backbones \cite{stilli2014shrinkable, yang2020geometric}, but this antagonistic mechanism can only regulate stiffness within a small range. Other similar approaches like air inflation \cite{pang2020coboskin}, particle jamming \cite{langer2018stiffening}, lever structure \cite{herzig2018variable}, heat sensitive smart materials \cite{zhang2023bioinspired} and patterned structures \cite{zhang2022novel, wang2019variable}, all suffered from long response time or bulky body or complicated activation steps. Most variable-stiffness designs could not selectively regulate directional stiffness, as the bending stiffness in deformation plane should be far lower than that in the lateral direction.
 
Tubular continuum robot characterizes central hollow space for tools and slender body. Oliver et. al. \cite{oliver2021concentric, oliver2017concentric} proposed a pair of asymmetrically cut tubes assembled at the tip to induce bending, where relative translation between the outer tube and the inner tube could deform the patterned section bidirectionally, as shown in Fig. \ref{fig:basics} (b). It differs from conventional concentric tube robots since two tube form a pair and the deformation was realized through antagonistic force. However, it also characterizes insufficient stiffness while the robot body is slender, which thereby requires delicate design on the patterned slits. Therefore, we also employed this category of tubular robot arm and enhance its overall stiffness though parameter-optimization design.

As shown in Fig. \ref{fig:basics} (b), relatively push/pull between coupled tubes deforms the CATR, while the parameters of the patterned slits is involved in the dexterity and stiffness. Patterned area poses great influence on stiffness, stability, maximum deformation angle, and more importantly acting force of the manipulator \cite{park2021design, barrientos2023asymmetric}. Most existing tubular continuum robots were patterned without theoretical support, such as triangular shaped \cite{wei2012modeling}, horizontal/vertical \cite{bell2012deflectable} and dog-bone shaped slits \cite{kim2019continuously}, lacking enough parametric optimization to the slit parameters and slit type selection. Additionally, only uncut area compensates the external disturbance, so that fatigue failure and buckling may present when tubular robot is in large deflection. To address the issue, this work employed tenon-mortise shaped slits, which could constrain the maximum deformation angle and enhance the axial stiffness with actuation power, as shown in Fig. \ref{fig:basics} (b). During selecting the slits parameters, bending stiffness in  lateral plane was considered, while maximizing the bending angle to enhance dexterity. 

Furthermore, investigating the kinematics and dynamics of the soft manipulator especially in large deflection status, is indispensable for accurate and dexterous manipulation. Various/constant curvature-based models assume the deformed backbone as an arch \cite{webster2010design}, so the kinematics is divided into shape-tip mapping and actuation configuration-shape space mapping, which did not take material properties into account; Statics approaches include Cosserat Rod theory \cite{tummers2023cosserat}, beam theory \cite{he2018research} and finite element method (FEM) \cite{singh2021dynamic}; dynamics were built on the basis of Newton-Euler equation \cite{jensen2022tractable}, Newton-Lagrange equation \cite{falkenhahn2014dynamic} and virtual work principle \cite{rone2013continuum}, where partial derivative equations should be solved. For CATR, the pure bending motion is a twofold discrete variable (discrete inherent direction angle is $0$ or $\pi$, and continuous bending angle), so its inverse kinematics consists of hybrid variables.
 
In this work, we proposed an  co-axial antagonistic tubular robot (CATR), corresponding parametrized slit design scheme, kineostatic modelling and control strategy for ambidextrous manipulation in mesco-scale space. Our contribution includes:\\
1) Parameterized tenon-mortise slit design on CATR for stiffness and dexterity enhancement.\\
2) Kineostatic modelling between actuation space and task space.\\
3) Optimization-based control strategy have been proposed for multi-segment CATR.\\
4) Experimental validation for the proposed methodology and potential application were demonstrated.

The rest of this work is organized as follows. In Section- \ref{Kine Model}, we elaborate on the basic working principle and modelling of the antagonistic co-axial tubular robot with a single segment.  Then, the parametrized design theory on the slits patterned along the steerable section was presented in Section-\ref{Design Opt Ana}. Analysis and control strategy for dual-segment design are detailed in Section-\ref{Modelling}. Experimental evaluation to demonstrate its performance and applicability is presented in Section-\ref{Experiments}. Finally, conclusion and future work are organized in Section-\ref{Conclusion}.
\section{Kinetostatic Modelling}\label{Kine Model}
As shown in Fig. \ref{fig:basics} (b), a single-segment CATR consists of two thin-walled metal tubes (generally NiTi and stainless steel). Each tube has a patterned section and they are fixed at the tip with a shift angle of $180^\circ$. The inner tube is co-axially arranged with the outer tube, and they are fixed at the tip. Due to the asymmetrical setting, axial compression (pulling) or tension (pushing) acting at the inner tube bends the steerable section upwards or downwards, respectively, which is also shown in Supplementary Video. Patterned slits were manufactured by high-precision laser cutting technology, and the two tubes are fixed by laser welding process. Because of the nested configuration, the inner tube always yields to the shape of the outer. Since relative pushing/pulling between the two tubes bends the steerable section, we quantify the bending DoF by pushing/pulling distance of the inner tube, while keeping the outer tube axially static.

In this work, the outer tube is assumed fixed and actuation space is characterized by the relative push/pull distance and push/pulling force. Kinetostatic mapping between actuation space to shape space is detailed in this section.
\subsection{Mapping between Push/Pull Distance and Bending Angle}\label{P-theta}
\begin{figure}
	\centering
	\includegraphics[width=1\linewidth]{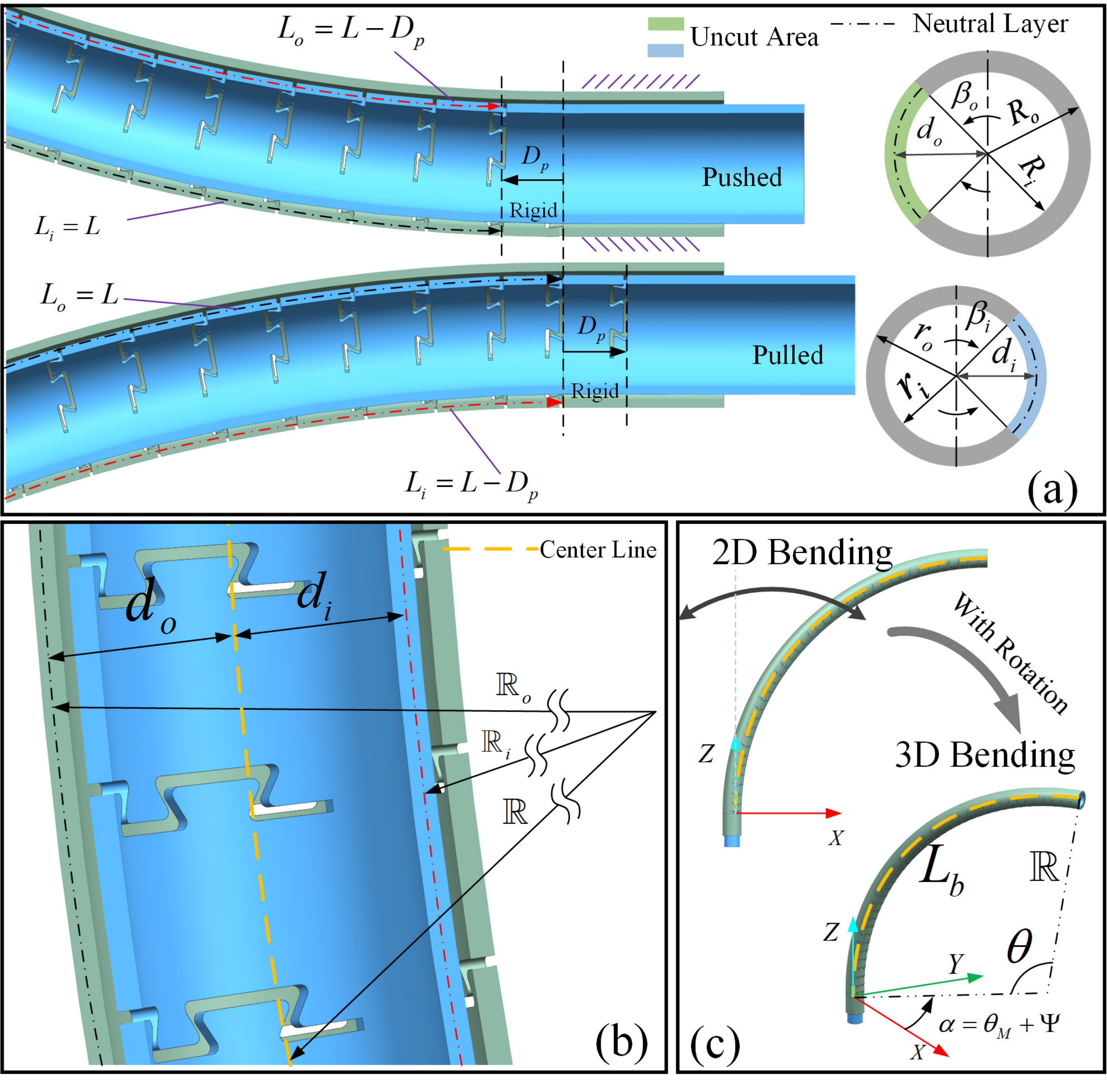}
	\caption{(a) Cross section of the robot arm under pushed and pulled status. (b) Bending parameters of the co-axial tubes. (c) Base rotation transfers bidirectional bending into 3D bending.} 
	\label{fig:geometry}
\end{figure}
The push/pull distance $ D_p $ (Push: $ D_p>0 $, Pull: $D_p<0$) of the inner tube is highly involved in the bending angle of the robot arm, as shown in Fig. \ref{fig:geometry} (a). The mapping between push/pull distance-bending angle is established under the following assumptions:\\
\textbf{Assumption 1:} As patterned slits are densely distributed ($ < 0.5mm$), the shape of the robot arm in bending status is deemed as a continuous arch with constant curvature.\\
\textbf{Assumption 2:} The gap between the two tubes is also neglected. The inner radius of the outer tube $R_i$ is just $ 0.05mm $ larger than the outer radius of the inner tube $ r_o $, and the two tubes are always co-axially configured.\\
\textbf{Assumption 3:} The uncut areas of the two tubes are exactly $ 180^\circ $ shifted and their relative axial rotation is constrained, generating only 2D bending.

For consistency, the patterned section of the two tubes are with the same length $ L $. Due to the outer tube's greater stiffness and geometric constraint of the inner tube, the shape of the inner tube tends to conform to that of the outer tube. For ease of illustrating bending actuation, the outer tube is static and the inner tube is pushed or pulled with respect to (w.r.t.) the outer. When the inner tube is being pulled, part of its patterned section (length of $ D_p $) is in the rigid section of the outer tube. On the contrary, when it is pushed, its rigid part (also length of $ D_p $) is inserted into the patterned section of the outer tube, as shown in Fig. \ref{fig:geometry} (a). Consequently, the actuation mode (pushed or pulled) determines the bendable section $ L_o $ (outer tube) and $ L_i $ (inner tube):
 \begin{equation}\label{Deformable_Section}
\left\{ {\begin{array}{*{20}{c}}
		{{L_o} = L,{L_i} = L - \left| {{D_p}} \right|}&{{\text{Pulled}}}\\
		{{L_i} = L,{L_o} = L - \left| {{D_p}} \right|}&{{\text{Pushed}}}
\end{array}} \right.
 \end{equation}

During bending, the length of the neutral layer of each tube maintains inextensible for both inner tube and the outer tube (see Fig. \ref{fig:geometry} (b)). The distance between the neutral layer and the center line is defined by $ d_o $($ d_i $):
\begin{equation}\label{Neutral_Layer}
	{d_o} = \frac{{2\int_{\frac{{\pi  - {\beta _o}}}{2}}^{\frac{{\pi  + {\beta _o}}}{2}} {\int_{{R_i}}^{{R_o}} {{r^2}\cos (\theta )drd\theta } } }}{{(R_o^2 - R_i^2){\beta _o}}}
\end{equation}
where $ R_o $($ R_i $) denotes outer(inner) diameter, and $ \beta_o $ is the central angle of the uncut area. This is identical for the inner tube, so deriving $ d_i $ is omitted for brevity. 

Thus, the bending radius of them are respectively derived as ${\mathbb{R}}_{j} = {{{L_j}} \mathord{\left/{\vphantom {{{L_o}} \theta }} \right.		\kern-\nulldelimiterspace} \theta }$ ($ \mathbb{R}_{j}=\mathbb{R}_i $ or $ \mathbb{R}_o $), where $ \theta $ is the bending angle illustrated in Fig. \ref{fig:geometry} (c). They share a common center of arch, such that:
\begin{equation}\label{Bending Angle}
	\left| {{\mathbb{R}}_{o}}-{{\mathbb{R}}_{i}} \right|={{d}_{o}}+{{d}_{i}}.
\end{equation}
Therefore, the bending angle is then obtained as $\theta  = \frac{{\left| {{D_p}} \right|}}{{{d_o} + {d_i}}}$. Geometrically, the backbone length $ L_b $ also depends on the bending direction and the bending angle:
\begin{equation}\label{Backbone Length}
	\left\{ \begin{matrix}
		{{L}_{b}}=\theta ({{\mathbb{R}}_{o}}+{{d}_{o}}) & \text{Pushed}  \\
		{{L}_{b}}=\theta ({{\mathbb{R}}_{i}}+{{d}_{i}}) & \text{Pulled}  \\
	\end{matrix} \right.
\end{equation}
To enable larger task space and 3D bending, a revolve DoF is attached at the CATR's base, parametrized as rotation angle $ \theta_M $ (see Fig. \ref{fig:geometry} (c)). Because of the bidirectional bending mode, pushing/pulling the inner tube essentially generates an inherent direction angle $\Psi $ ($\Psi=0 $ or $\pi$), as shown in Fig. \ref{fig:basics} (b). So, the combined direction angle is $\alpha=\theta_M+\Psi$, and the tip position $ P(x,y,z) $ could be obtained:
\begin{equation}\label{Tip Position}
	\left[ \begin{matrix}
		x  \\
		y  \\
		z  \\
	\end{matrix} \right]=\frac{{{L}_{b}}}{\theta }\left[ \begin{matrix}
		(1-\cos \theta )\cos \alpha   \\
		(1-\cos \theta )\sin \alpha   \\
		\sin \theta   \\
	\end{matrix} \right] 
\end{equation}
\subsection{Statics Modeling}
\begin{figure}
	\centering
	\includegraphics[width=1\linewidth]{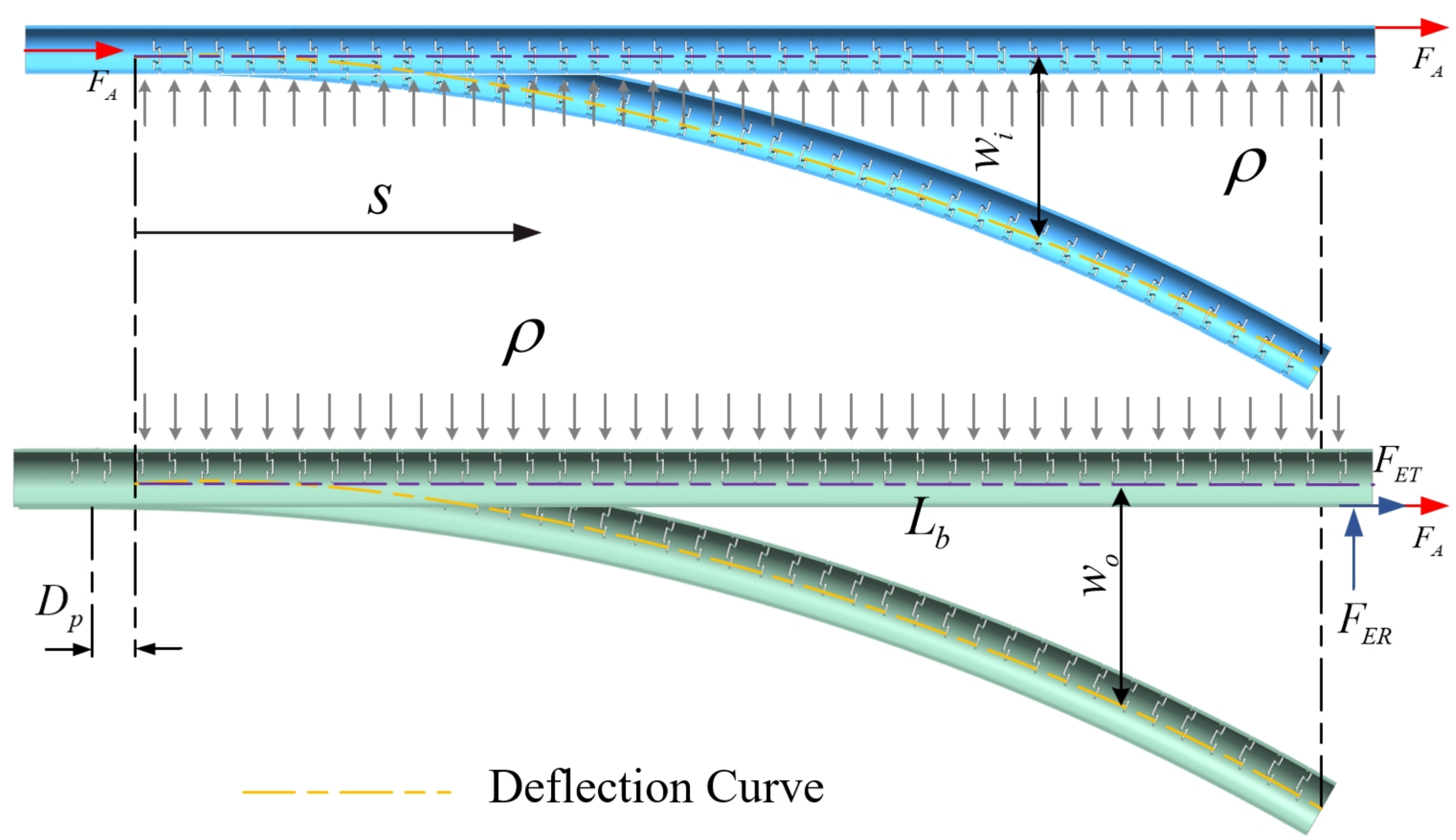}
	\caption{Statics model of the tubes. The inner tube is only activated by actuation force and the distributed load $ q $ from the outer tube. Apart from this, the outer tube is also excited by tangent external force $ F_{ET} $ and radial external force $ F_{ER} $. }
	\label{fig:statics-model}
\end{figure}
In addition to push/pull distance, actuation force $ F_A $ (push/pull force) acting along the axial direction of the inner tube is also an important index to characterize the bending angle. The moments at the tubes' tip simultaneously deflect them. As shown in Fig. \ref{fig:statics-model}, the inner tube is only excited by actuation force $ F_A $ and the interaction load from the outer tube, while the outer one is both excited by actuation moment and external forces. The statics model is established under following assumptions:\\
\textbf{Assumption 4:} The interaction force between the inner tube and the outer tube is assumed as distributed load  $\rho $ at the surface. Friction between the both tubes are neglected.\\
\textbf{Assumption 5:} External forces only acts on the outer tube's tip, i.e., radial force $ F_{ER} $ and tangent force $ F_{ET} $. \\
\textbf{Assumption 6:} Shear strain is neglected for the thin and slender robot arm. 
The moment $ M(s) $ at the cross section $ s $ is defined via Euler-Bernoulli theory as:
\begin{equation}\label{Moment}
	\begin{aligned}
	{M_i}(s) = &\int_s^{{L_s}} {\rho({L_s} - x)dx - {F_A}{d_i}} \\
	{M_o}(s) =  &- \int_s^{{L_s}} {\rho({L_s} - x)dx + ({F_A} + {F_{ET}}){d_o}}\\
	& + {F_{ER}}({L_s} - s)
	\end{aligned}
\end{equation}
where only bending backbone length $ L_s $ and actuation force $ F_A $ are known. With Euler-Bernoulli beam theory, the bending curvatures $ {w''_i}(s), {w''_o}(s) $ of the tubes are established:
\begin{equation}\label{deflection}
	\frac{{M_i(s)}}{{E{I_i}}} = {w''_i}(s),\frac{{M_o(s)}}{{E{I_o}}} = {w''_o}(s)
\end{equation}
where $ w(s) $ denotes the deflection curve, $ E $ is the Young's Elasticity modulus, and $ I $ is the second moment of inertia. Since the tubes are always constrained to keep co-axial, their deflection curves are deemed as identical, so $ w(s)=w_i(s)=w_o(s) $. Then the problem is converted to solve ordinary derivative equations to obtain the deflection curve function, which requires boundary conditions: the deflection, deflection angle, and curvature at the position $ s=0 $ are zero. 
\begin{equation}
w(0) = w'(0) = w''(0) = 0
\end{equation}
This is enough to solve the deflection curve on the condition that $F_{ET}=F_{ER}=0$. If there is external excitation, i.e. ${F_{ER}} \ne 0, {F_{ET}} \ne 0$, more boundary conditions could be added to solve the deflection curve:
\begin{equation}\label{ODE_CON}
\left\{ {\begin{array}{*{20}{c}}
		{{w_i}(s = {{{L_b}} \mathord{\left/
					{\vphantom {{{L_b}} 2}} \right.
					\kern-\nulldelimiterspace} 2}) = {w_o}(s = {{{L_b}} \mathord{\left/
					{\vphantom {{{L_b}} 2}} \right.
					\kern-\nulldelimiterspace} 2})}\\
		{{{w'}_i}(s = {{{L_b}} \mathord{\left/
					{\vphantom {{{L_b}} 2}} \right.
					\kern-\nulldelimiterspace} 2}) = {{w'}_o}(s = {{{L_b}} \mathord{\left/
					{\vphantom {{{L_b}} 2}} \right.
					\kern-\nulldelimiterspace} 2})}
\end{array}} \right.
\end{equation}

\begin{figure}
	\centering
	\includegraphics[width=1\linewidth]{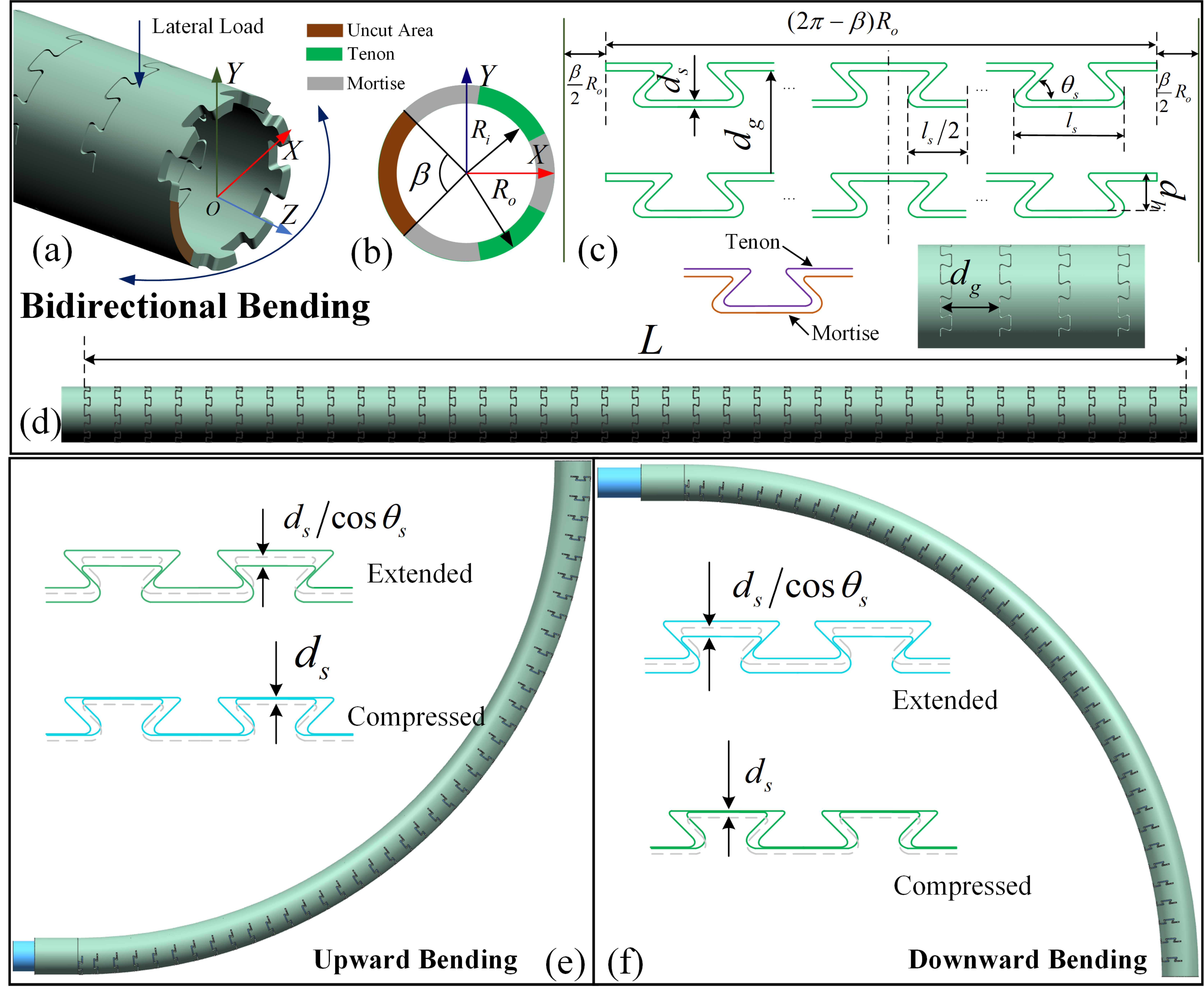}
	\caption{Asymmetrically patterned tube with mortise-tendon slits. (a) 3D sectional view of the mortise-tendon slits. (b) Sectional view. (c) Expanding view of the patterned tube with parameters of the slits. (e) and (f) tenon-mortise status of inner tube and outer tube in upward bending and downward bending conditions.}
	\label{fig:slitparaillu}
\end{figure}

\section{Design Optimization and Analysis}\label{Design Opt Ana}
While the geometry-based approach and beam theory have been utilized to model a single-segment CATR arm, the design parameters of slits are involved in the maximum bending angle and stiffness. In this section, the method to solve the optimal design parameters of the patterned slits is detailed.
\subsection{Parametrized Slit Design}
Inspired by traditional Chinese carpentry, we have employed tenon-mortise slits patterned along the tubes to form the steerable section. As shown in Fig. \ref{fig:basics} (b), the outer tube's slits are compressed and extended respectively while bending upwards and downwards, vice versa for the inner tube. One slit has tenons and mortises, and they are buckled together while extended, reaching the maximum bending angle. Therefore, to enhance the dexterity and maneuverability, we desire to enhance the maximum bending angle. Simultaneously, the tubular robot arm should be stiff to offset external force, which is then transferred into a constrained optimization problem.

For a raw tube with basic geometric parameters including Young's Elastic Modulus $ E $, inner diameter $ R_i $, outer diameter $ R_o $ and steerable section length $L$, the to-be-solved parameters are central angle $\beta $ representing the uncut area in cross section, length of tenon $ l_s $, gap distance $ d_g $ between adjacent slits, number of slits $N$, number of tenon-mortise combination $n$ in one slit, slit width $d_s$, height of slits $ d_h $ and tilted angle $ \theta_s $ of the tendon, as shown in Fig. \ref{fig:slitparaillu} (a) and (b). Let $S = \{ L,\beta ,{l_s},{d_g},N,n,{d_s},d_h,{\theta _s}\} $ denote the to-be-optimized parameters. $S^o$ and $S^i$ respectively represent the parameters for the outer tube and the inner tube. In the following, the subscript or superscript $o$ and $i$ respectively denote the parameter for the outer and the inner tube, and equations without subscript/superscript are applicable for both.
\begin{table}[h]
	\caption{Minimal and maximum limitations of each parameter. 'Constant' means the parameter is manually predefined.}
	\label{tab:Parameters Limitation}
	\centering
	\begin{tabular}{p{1.5cm}<{\raggedright}  p{0.6cm}<{\raggedright}   p{0.6cm}<{\centering} p{1.7cm}<{\raggedright}  p{0.5cm}<{\centering}   p{0.5cm}<{\centering} }
		\hline
		Variable & {Min.}      & \multicolumn{1}{c|}{Max.}      & Variable & Min. & Max. \\ \hline
		$L$ (mm)          & \multicolumn{2}{c|}{Constant} & $N$ (integer)     & 1       & 1000    \\
		$R_i$/$r_i$ (mm)    & \multicolumn{2}{c|}{Constant} & $n$ (odd num.)    & 1       & 5       \\
		$R_o$/$r_o$ (mm)    & \multicolumn{2}{c|}{Constant} & $d_s$ (mm)         & 0.03    & 0.06    \\
		$\beta$ (rad)      & 0             & \multicolumn{1}{c|}{$2\pi$}           & $d_h$ (mm)         & 0.25    & 0.3     \\
		$l_s$ (mm)         & 0.1           & \multicolumn{1}{c|}{0.7}          & $\theta_s$ ($^\circ$)     & 25      & 60      \\
		$d_g$ (mm)         & 0.3           & \multicolumn{1}{c|}{0.6}         &                 &         &         \\ \hline
	\end{tabular}
\end{table}

\textbf{(1) Design Goals} Our aim is to enhance dexterity, so the mathematics goal is maximizing the bending curvature. Simultaneously, the overall stiffness should be maximized, which is the other goal. The first goal is parametrized by the maximum curvature $\kappa_M$:
\begin{equation}
	{\kappa _M} = \frac{{{\theta _M}}}{L},
\end{equation}
where $\theta_M$ is the maximum bending angle. With the bidirectional bending, $\theta_M$ has two folds, i.e. $ \theta _o^D = {\theta _M}$ for the downward bending and $ \theta _o^U = {\theta _M}$ for the upward bending:
\begin{equation}
{\theta _M} = \left\{ {\begin{array}{*{20}{l}}
		{\theta _o^U = \frac{{{N_o}d_s^o}}{{({d_o} + {d_i})\cos \theta _s^o}}}&{\text{Pulled}}\\
		{\theta _o^D = \frac{{{N_o}d_s^o}}{{({d_o} + {d_i})}}}&{\text{Pushed}}
\end{array}} \right.
\end{equation}
Here, we select $ \theta _o^U = {\theta _M}$ to organize the objective function because when tenons and mortises are both buckled together, the stiffness is enhanced. In addition, the outer tube has larger $I_o$, so the stiffness in buckled status is higher. For the inner tube, we require $\theta _o^U = \theta _i^U$, ensuring bending angles of the two tubes simultaneously reach the maximum. As a result, while all the tenon-mortise structures of the both tubes are simultaneously buckled together, and the overall stiffness reached the maximum, as shown in Fig. \ref{fig:slitparaillu} (e) and (f).

 Additionally, the performance to against lateral loads also matters, as illustrated in Fig. \ref{fig:slitparaillu} (a), so the stiffness in $YOZ$ plane are considered for optimization. The stiffness is generally characterized by: $EI=E(I_i+I_o)$, and $I$ is the second moment of inertia involved in the design parameters:
 \begin{equation}
 {I_o} = \int_A {{y^2}} dA = \int_{{R_i}}^{{R_o}} {\int_{ - \frac{{{\beta _o}}}{2}}^{\frac{{{\beta _o}}}{2}} {{{(r\sin \theta )}^2}rd\theta dr} } 
 \end{equation}
 which is similar for the inner tube and omitted. Therefore, the optimization goal is defined as:
 \begin{equation}
\min .(\underbrace {\frac{1}{{{\kappa _M}}}}_{{f_1}},\underbrace {\frac{1}{{{I_i} + {I_o}}}}_{{f_2}})
 \end{equation}
\textbf{(2) Constraints} Generally, the tube diameters were set following the actual application, such as the size of natural orifices for MIS. Basically, the parameters should fall within a given range:
\begin{equation}
	{S_{\min }} \le S \le {S_{\max }}
\end{equation}

The boundary value of each parameter is listed in TAB. \ref{tab:Parameters Limitation}. In this work, we define the outer diameter, inner diameter, and length of steerable section manually, which are mutually independent. While, some parameters are mutually coupled:
\begin{equation}
\left\{ {\begin{array}{*{20}{l}}
		{(2{l_s} - \frac{{2{d_h}}}{{\tan {\theta _s}}})n = (2\pi  - \beta ){R_o}}\\
		{N({d_s} + {d_g}) = L}
\end{array}} \right.
\end{equation}
where the first equation presents the relationship between tenon-mortise structure and the patterned length in one slit, and the second equation denotes the relationship between slit number and deformable section length.

With the requirement $\theta _o^U = \theta _i^D$, the slit parameters of the two tubes are also constrained:
\begin{equation}
	N_o\frac{{d_s^o}}{{\cos \theta _s^o}} =N_i d_s^i
\end{equation}
\textbf{(3) Solution Approach} 
Since the two objective functions $f_1$ and $f_2$ have totally different dimensions, weighted sum could not form a single-objective problem. An parameter set could not be found to simultaneously reach the minimal values of $f_1$ and $f_2$. Therefore, Pareto Optimal Solution approach is employed  \cite{mattson2004smart}. First, the solution set $S_{1m}$ was searched to minimize $f_1$ and got a minimum $f_{1m}$, with $f_2$ ignored. Similarly, another set $S_{2m}$ was sought to map with $f_{2m}$ while $f_1$ was ignored. Thus, the set $ S_{1m}\le S \le S_{2m}$ forms a Pareto frontier. 
\begin{algorithm}[t]
	\caption{Optimal Design Solution Approach}
	\label{alg::Design Algorithm}
	\KwIn{Design Domain Boundaries $S_{\min}$ and $S_{\max}$. Initialize  $\epsilon$ ($> 1$).}
	\KwOut{Optimal Design Parameters $S_{op}$}
	\BlankLine
	Stage 1: \\
	Solving Individual Optimum of $f_1$ and $f_2$\\
	Initialize $S_0$ randomly\\
	Calculate the optimum of $f_1$ and $f_2$ while ignoring the other\\
	$\left\{ {\begin{array}{*{20}{c}}
			{{s_{1m}} = \arg \min {f_1}(S)}\\
			{{\rm{w}}{\rm{.r}}{\rm{.t}} {\rm{. Constraints}}}
	\end{array}} \right.{\rm{and}}\left\{ {\begin{array}{*{20}{c}}
			{{s_{2m}} = \arg \min {f_2}(S)}\\
			{{\rm{w}}{\rm{.r}}{\rm{.t }} {\rm{. Constraints}}}
	\end{array}} \right.$\\
	
	Get:\\
	Pareto Frontier: $ S_{1m}\le S \le S_{2m}$\\ 
	$f_{1m}=f_1(S_{1m})$, $f_{2m}=f_1(S_{2m})$
	\BlankLine
	Stage 2: \\
	Minimize $f_1$ while pursing a minimum $f_2(S)$\\
	\While{$f_1$ not in convergence or $i \le i_{\max}$}{
		$\begin{array}{l}
			{S_{op}} = \arg \min {f_1}(S)\\
			{\rm{w}}{\rm{.r}}{\rm{.t}}{\rm{.}}\left\{ {\begin{array}{*{20}{c}}
					{{f_2}(S) \le \epsilon {f_{2m}}}\\
					{{\rm{Constraints}}}
			\end{array}} \right.
		\end{array}$\\
		$\epsilon  \leftarrow \epsilon  - 0.01$\\
		$i \leftarrow (i+1) $
	}
	
\end{algorithm}

Then, the bending curvature is separately deemed as the objective function, and stiffness function tries to reach the minimum, where the frontier $f_{2m}$ is leveraged as another constraint. The algorithm is clarified in Algorithm \ref{alg::Design Algorithm}. An variable factor $\epsilon >1$ is introduced to indirectly form another constraint in optimization. Initially, $\epsilon$ is large so that the $S_{op}$ could easily satisfy the newly constructed constraints. Then, the $\epsilon$ begins to decrease by $0.01$ and the optimization runs iteratively. The whole optimization process stops until the objective function $f_1(S)$ converges or the iteration step reaches to the predefined number $i_{\max}$. This is executed by fmincon() in MATLAB.

\begin{figure*}
	\centering
	\includegraphics[width=1\linewidth]{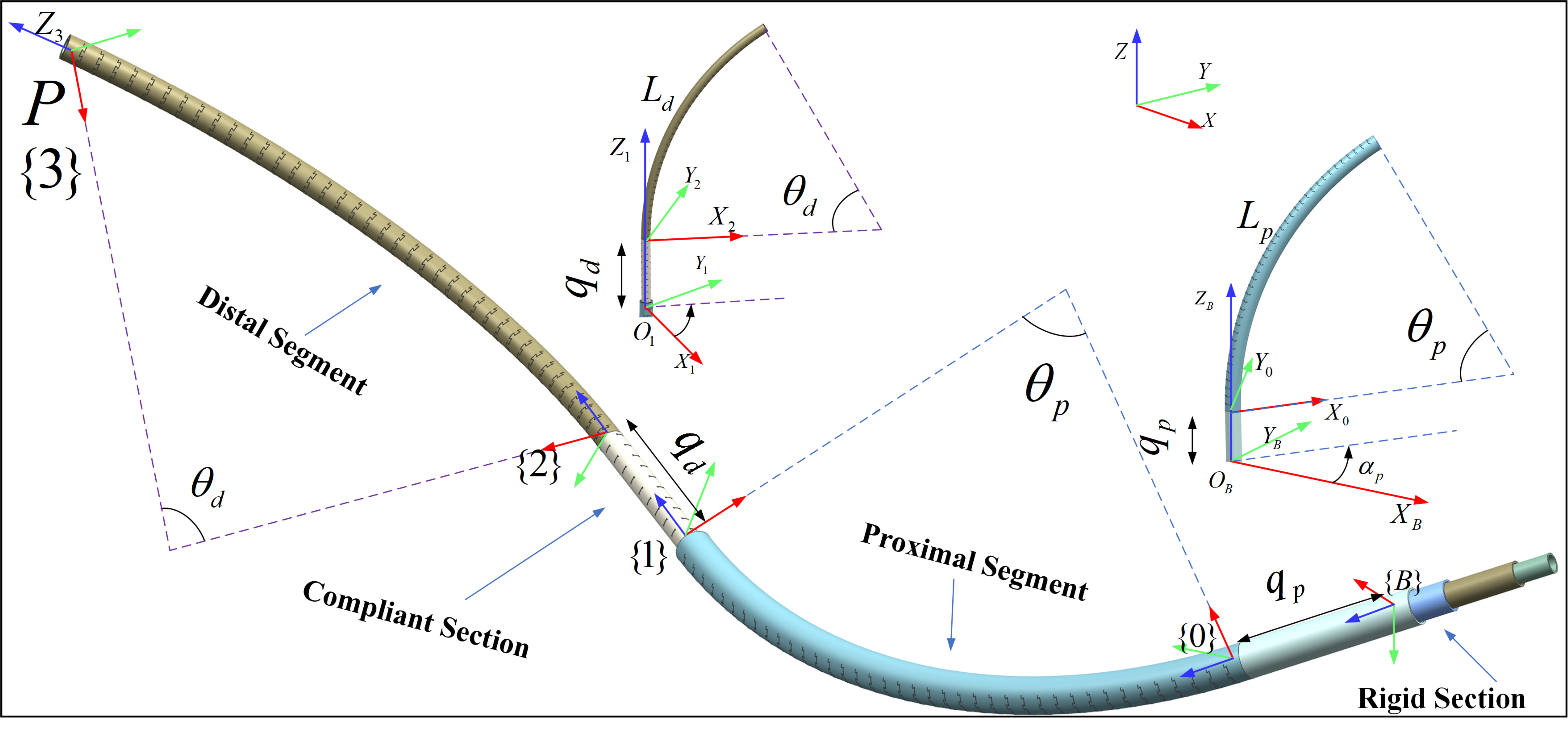}
	\caption{Dual-segment CATR. The distal segment passes through the hollow space of the proximal segment, and then the robot arm has totally 6 DoFs.}
	\label{fig:dualsegk}
\end{figure*}

\section{Multi-Segment Design and Modelling}\label{Modelling}
The tip position of a single segment is derived from \eqref{Tip Position}, from which the three components ($x,y,z$) are coupled with bending angle $\theta$ and direction angle $\alpha$. The orientation is coupled with position, so the dexterity and task space are marginal. The thin-walled structure is advantageous for configuring another thinner CATR to form a dual-segment slender robot arm, as shown in Fig. \ref{fig:dualsegk}. Consequently, the tip orientation is not tightly coupled with the position, so the mapping between actuation inputs and the tip pose should be further established for accurate manipulation.
\subsection{Kinematics}
Fig. \ref{fig:dualsegk} shows the shape and defined coordinate frames of a dual-segment CATR. The forward kinematics model was built by Homogeneous Transformation Matrix, and the tip point $P \in {R^6}$
w.r.t. the base frame $\{B\}$ is:
\begin{equation}
	\begin{array}{l}
		\begin{aligned}
			{}_B^3T = {T_z}({q_p}){R_z}({\alpha _p}){T_x}(\frac{{{L_p}}}{{{\theta _p}}}){R_y}({\theta _p}){T_x}( - \frac{{{L_p}}}{{{\theta _p}}})\\
			{T_z}({q_d}){R_z}({\alpha _d}){T_x}(\frac{{{L_d}}}{{{\theta _d}}}){R_y}({\theta _d}){T_x}( - \frac{{{L_d}}}{{{\theta _d}}})
		\end{aligned}	
	\end{array}
\end{equation}
where $T_i$ and $R_i$ respectively denote translation and rotation matrix about axis $i$. $q$ is translation distance, $\alpha$ is rotation angle, and $L$ is backbone length (subscript $p$ and $d$ respectively stands for proximal segment and the distal). Each shape parameter is illustrated in Fig. \ref{fig:dualsegk}. Then, the tip pose $P(x,y,z,{\tau _x},{\tau _y},{\tau _z})$ ($\tau$ is Euler angle) could be accordingly solved by ${}_B^3T$. With the mapping between push/pull distance and bending angle, which is detailed in Section \ref{P-theta}, forward kinematics is obtained.

Inverse kinematics is also indispensable to bring robot tip to a given desired task configuration $\tilde P$. Actuation inputs $\tilde{A}$ to map with $\tilde{P}$ should be quickly solved and commanded to actuators. Because of hyper redundancy, i.e., multiple actuation inputs could map with a $\tilde{P}$, directly solving $\tilde{A}$ is impossible. In addition, pushing/pulling mode also poses discrete increment in direction angle $\alpha$:
\begin{equation}\label{Rot_Angle}
\alpha  = \left\{ {\begin{array}{*{20}{c}}
		{{\theta _M}}&{{\rm{Pushed}}}\\
		{{\theta _M} + \pi }&{{\rm{Pulled}}}
\end{array}} \right.
\end{equation}
where $\theta_M$ denotes the rotation DoF. For ease of mathematical expression, \eqref{Rot_Angle} is revised as: 
\begin{equation}
	\alpha  = {\theta _M} + (\frac{{1 + \emph{sign}({D_p})}}{2})\pi 
\end{equation}
where $\emph{sign}(\cdot)$ is symbolic function.

The distal segment could also partially retract into the hollow space of the proximal segment, i.e. $q_d<0$, and now the direction angle of the distal segment should keep identical with the proximal. Otherwise, the two segments are not able to bend, so we require that:
\begin{equation}
	\left\{ {\begin{array}{*{20}{c}}
			{{\alpha _d} = {\alpha _p}}&{{q_d} < 0}\\
			{{\alpha _d} \in \left[ {0,2\pi } \right]}&{{q_d} \ge 0}
	\end{array}} \right.
\end{equation}
Similarly, it is modified for ease of calculation:
\begin{equation}
	{\alpha _d} = {\alpha _p}(\frac{{1 - \emph{sign}({q_d})}}{2}) + \theta _{_M}^d + (\frac{{1 + \emph{sign}(D_p^d)}}{2})\pi 
\end{equation}
where $\theta _{_M}^d$ and $D_p^d$ respectively denote rotation angle and push/pull distance of the distal segment. After analyzing the mutually coupled factors, we should solve all of them  $\tilde A({q_p},D_p^p,\theta _M^p,{q_d},D_p^d,\theta _M^d)$ to minimize $\left\| {FK(\tilde A) - \tilde P} \right\|$, where $FK(\cdot)$ is forward kinematics derived above.

Then, the inverse kinematics model is converted to a constrained optimization problem:
\begin{equation}\label{IK_Con_Op}
\begin{array}{l}
	\tilde A = \text{argmin} \left\| {FK(A) - \tilde P} \right\|\\
	s.t.\left\{ {\begin{array}{*{20}{l}}
			{{q_{p,\max }} \ge {q_p} \ge 0,{q_{d,\max }} \ge {q_d} \ge  - {L_p}}\\
			{2\pi  \ge \theta _M^p,\theta _M^d \ge 0}\\
			{{D_{\max }} \ge D_p^p,D_p^d \ge {D_{\min }}}
	\end{array}} \right.
\end{array}
\end{equation}
where the constraint limitations are set based on robot physical characteristics. Since direction angle $\alpha$ is jointly determined by rotation DoF $\theta_M$ and rotation motion is smooth during motion, we pursue minimal variation of push/pull distance between adjacent control instances to avoid snap. Therefore, the objective function in \eqref{IK_Con_Op} is updated as:
\begin{equation}
\begin{array}{l}
	\begin{aligned}
	\tilde A(k) =& \text{argmin} \{ \left\| {FK(A(k)) - \tilde P(k)} \right\|\\
& + \omega_D \left\| {D_p^p(k) - D_p^p(k - 1) + D_p^d(k) - D_p^d(k - 1)} \right\|\} 
	\end{aligned}
\end{array}
\end{equation}
where $k$ denotes the $k\text{th}$ control instance, and $\omega_D=0.1$ is a weighting coefficient.

In terms of solving the constrained optimization problem, we used Genetic Algorithm (GA), and one of its notable advantages is insensitive to gradient. In this work, actuation inputs consist of discrete variable, i.e. push or pull action, expressed by $\emph{sign}(\cdot)$. To process the searching process, the given desired tip task configuration is simplified into 4 elements: ${}^BP = ({}^Bx,{}^By,{}^Bz,\overrightarrow {{O_3}{Z_3}} )$, in which $\overrightarrow {{O_3}{Z_3}} $ is pointing direction of the end effector (see Fig. \ref{fig:dualsegk}). This is reasonable because target position and axial direction are two key control objectives in application.

\begin{figure*}[htbp]
	\centering
	\includegraphics[width=1\linewidth]{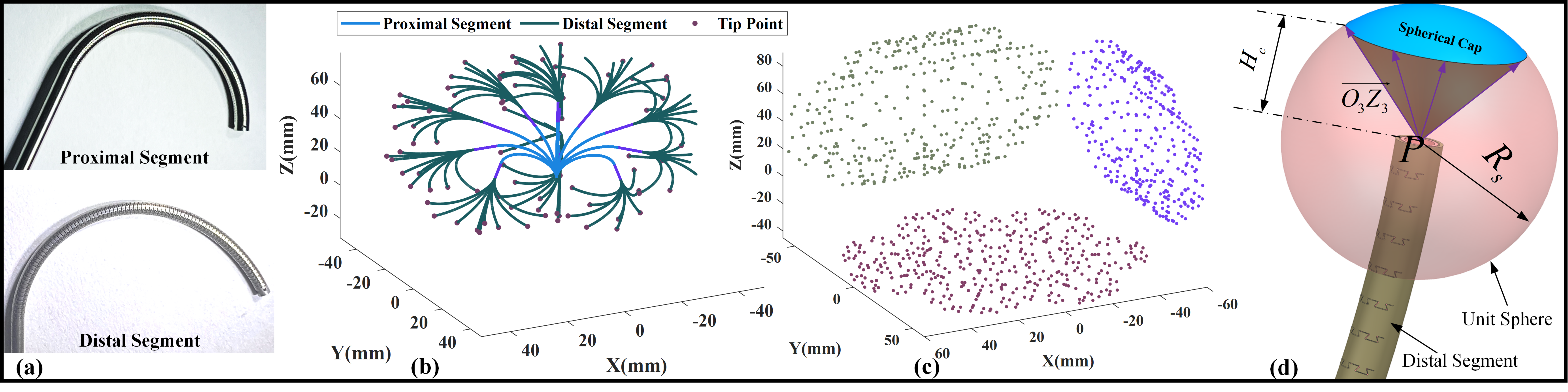}
	\caption{(a) Snapshots of proximal segment and distal segment in the maximum bending angle status. (b) Illustration of the dual-segment CATR. (c) Task space of tip point projected to 2D planes. (d) Tip dexterity characterized by a unit sphere and spherical cap.}
	\label{fig:taskspace}
\end{figure*}

\subsection{Analysis}
Reachability and dexterity are two crucial indexes to evaluate a robot's performance. In continuum robotics, reachability is generally quantified by the volume of tip's task space. Dexterity is characterized by the orientation range at a task point. The larger the bending angles, the more dexterous the robot arm. 
\begin{table}[h]
	\caption{Design parameters of proximal and distal segments.}
	\centering
	\label{tab:Geometric Parameters}
	\begin{tabular}{lllr}
		\hline
		\multicolumn{2}{c}{\textbf{Proximal Segment}}    & \multicolumn{2}{c}{\textbf{Distal Segment}} \\ \hline
		\multicolumn{2}{c}{Steerable Section Length (L)} & \multicolumn{2}{c}{30mm}                    \\
		$R_o^p$           & \multicolumn{1}{r|}{1.7mm}    & $R_o^d$                   & 1.5mm             \\
		$R_i^p$            & \multicolumn{1}{r|}{1.4mm}    & $R_i^d$                   & 1.2mm             \\
		$r_o^p$            & \multicolumn{1}{r|}{1.1mm}    & $r_o^d$                   & 0.9mm             \\
		$r_i^p$            & \multicolumn{1}{r|}{0.8mm}    & $r_i^d$                   & 0.6mm             \\
		$\theta _{\max }^p$     & \multicolumn{1}{r|}{$170^\circ$}      & $\theta _{\max }^d$             & $160^\circ$\\
		  	& \multicolumn{1}{l|}{}         & Compliant Section      & 40mm       \\ \hline           
	\end{tabular}
\end{table}

First, we analyze the reachability of the tip, and basic geometric parameters of robot is listed in TAB. \ref{tab:Geometric Parameters}, which is a typical example to show reachability. In TAB. \ref{tab:Geometric Parameters}, subscript $p$ and $d$ denote the parameter of proximal and distal segment, respectively. For example, $R_o^p$ indicates the outer diameter of outer tube of the proximal segment, and $R_i^p$ is its inner diameter. The distal segment has a special passive compliant section, whose shape is determined by the proximal segment, so the distal segment can extend maximally $10mm$, i.e., $q_{\max}^d=10$. Fig. \ref{fig:taskspace} (a) shows the snapshots of the two segments in maximum bending angle status, and the maximum bending angles of them are around $170^\circ$ and $160^\circ$, respectively. By setting discrete actuation inputs and sampling the tip position, we obtained the task space, as shown in Fig. \ref{fig:taskspace} (b). The translation DoF of the proximal segment $q_p$ was not activated, which only expands task space vertically. As shown in Fig. \ref{fig:taskspace} (c), the volume of task space is around $870 cm^3$. Notably, the tip could reach the area where $z$ value is negative, which further demonstrates the large task space and high dexterity.

Fig. \ref{fig:taskspace} (d) shows the proposed method to characterize the dexterity. The robot redundancy enables the tip reach a desired position and present different orientations. Purple arrows in Fig. \ref{fig:taskspace} (d) illustrates multiple $\overrightarrow {{O_3}{Z_3}}$, and all the reachable orientation forms a cone-shaped area. The larger the area of spherical cap, the more dexterous the robot. Then, dexterity could be mathematically expressed by the ratio between the surface area of spherical cap and the whole unit sphere's surface area:
  \begin{equation}
  	{\rm{Dex}} = \frac{{2\pi {R_s}{H_c}}}{{4\pi R_s^2}}
  \end{equation}
where $R_s=1$ is the radius of unit sphere, and $H_c$ is the distance between sphere center and the bottom of spherical cap, as depicted in Fig. \ref{fig:taskspace} (d).
\section{Experimental Validation}\label{Experiments}
In this section, the proposed optimization method was first validated, including the optimal design approach and inverse kinematics model, and tune their parameters for further evaluation in robot arm experimental platform. We have also built a compact actuation unit for the dexterous robot arm, with which some basic performance has been evaluated and its potential in medical application was demonstrated. Supplementary video shows all the motion of the robot.
\subsection{Tube Design, Fabrication and Actuation Unit}
First, the robot arm was designed and fabricated. The slits parameters along each tube were obtained following the design scheme. After assembly, the bending angle and stiffness of the two segments should be tested to validate the design scheme. Some basic geometric parameters of each tube are listed in TAB. \ref{tab:Geometric Parameters}. As biopsy gripper actuation wire and endoscopic camera wire (generally 0.5mm in diameter) should pass through the hollow space, $r_i^d$ was set to 0.6mm and the radial gap between tubes should be 0.05mm at least, so the outer diameter of proximal segment's outer tube $R_o^p$ was 1.7mm. Through the slits optimization algorithm, the parameters determining the steerable section are listed in TAB. \ref{tab: Optimal_Parameters_Slits}. As a result, the maximum bending angles of the proximal segment and the distal segment were $173^\circ$ and $162^\circ$, and Fig. \ref{fig:taskspace} (a) shows the snapshots, which is almost consistent with the theoretical value. In TAB. \ref{tab: Optimal_Parameters_Slits}, $N$ for each segment are both rounded to 60 and 74.

\begin{table}[h]
	\caption{The optimal parameters of slits patterned on tubes}
	\label{tab: Optimal_Parameters_Slits}
	\begin{tabular}{p{1.5cm}<{\raggedright} p{1.3cm}<{\centering} p{1.3cm}<{\centering} p{1.3cm}<{\centering} p{1.3cm}<{\centering}}
		\hline
		\textbf{}                 & \multicolumn{2}{c}{\textbf{Proximal Segment}}       & \multicolumn{2}{c}{\textbf{Distal Segment}} \\ \hline
		\textbf{Design Parameter} & \textbf{Outer Tube} & \textbf{Inner Tube} & \textbf{Outer Tube}  & \textbf {Inner Tube} \\ \hline
		$\beta$ (rad)                & 0.5070              & 0.5                  & 0.513                & 0.4992      \\
		$l_s$ (mm)                   & 0.6348              & 0.5815               & 0.6857               & 0.5775      \\
		$d_g$ (mm)                   & 0.4697              & 0.35                 & 0.4665               & 0.35        \\
		$N$                         & 59.8                & 73.17                & 59.94                & 73.19       \\
		$n$                         & 5                   & 5                    & 3                    & 3           \\
		$d_s$ (mm)                   & 0.0317              & 0.06                 & 0.0339               & 0.06        \\
		$d_h$ (mm)                   & 0.3                 & 0.25                 & 0.3                  & 0.25        \\
		$\theta$ (rad)                & 1.1238              & 0.9557               & 1.0893               & 0.9161      \\ \hline
	\end{tabular}
\end{table}
\begin{figure*}
	\centering
	\includegraphics[width=1\linewidth]{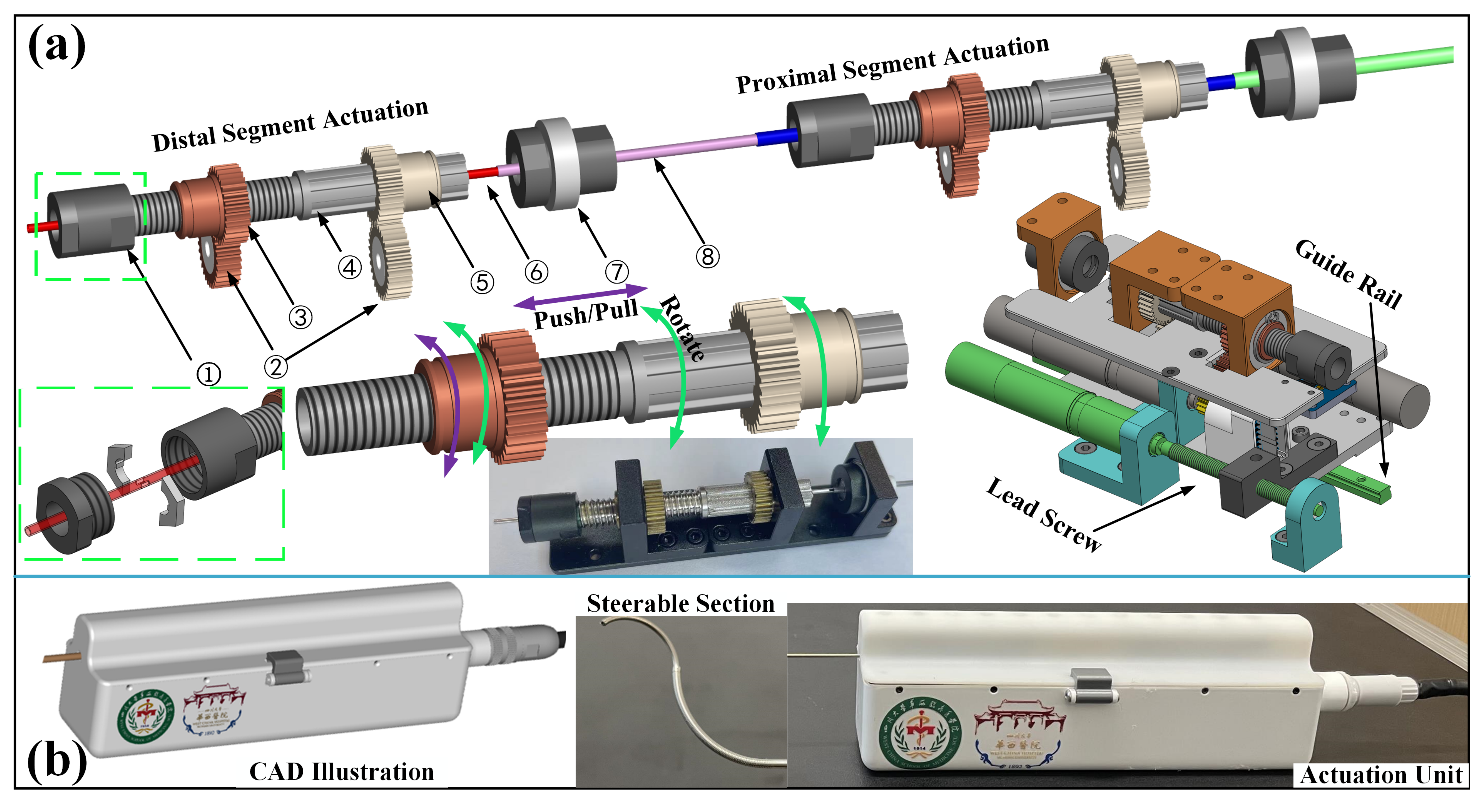}
	\caption{Compact actuator design. (a) Basic working principle of actuators. \textcircled{\scriptsize{1}}-Inner Tube Fixator, \textcircled{\scriptsize{2}}-Transmission Gears, \textcircled{\scriptsize{3}}-Thread Gear, \textcircled{\scriptsize{4}}-Spline-Thread Shaft, \textcircled{\scriptsize{5}}-Spline Gear, \textcircled{\scriptsize{6}}-Inner Tube, \textcircled{\scriptsize{7}}-Outer Tube Fixator, \textcircled{\scriptsize{8}}-Outer Tube. (b) CAD illustration and snapshot of the whole actuation unit.}
	\label{fig:actuation-unit}
\end{figure*}

In terms of manufacturing, raw stainless steel tubes (material: 304 stainless steel) were prepared to conduct high-precision laser cutting process. Next, the processed tubes were sent into ultrasonic cleaner to remove the debris trapped in the slits. Finally, a pair of nested tubes are fixed together by laser welding technology. The snapshots of the three machines are shown in supplementary material. The assembled segments could be steered manually, as shown in supplementary video.

For actuation, each segment requires three actuators, i.e., translation, rotation and push/pull DoFs. Essentially, the push/pull actuator also provides translation DoF, but the push/pull distance is small ($ - 2.5 \le {D_p} \le 2.5$). Therefore, we employed spline-thread shaft mechanism to generate bending and rotation. As shown in Fig. \ref{fig:actuation-unit} (a), the inner tube is fixed with the hollow shaft at the threaded section using threadlocker, and the outer tube is fixed on a bearing structure. Two gears with internal thread and internal spline were respectively mounted at the threaded section and the spline section of the shaft. The gears are individually controlled by two independent transmission gears, and they only provide rotation motion. While the two gears rotate with same speed, the shaft also rotate with the gears; While the spline gear maintain static and only the thread gear rotates, the shaft moves along the axial direction. Consequently, independent and synchronized control towards the two gears could easily control the push/pull distance $D_p$ and direction angle $\alpha$. The whole spline-thread module was mounted on a lead-screw module, so that a single segment CATR could be actuated by a compact unit. For the dual-segment design, two modules are linearly arranged, and the whole actuation unit is compact. Fig. \ref{fig:actuation-unit} (b) shows the CAD illustration and physical snapshot of the actuation unit. It is lightweight and can be handheld. Through CANBUS communication mechanism, user could control the steerable section at a remote console. 
\begin{figure*}
	\centering
	\includegraphics[width=1\linewidth]{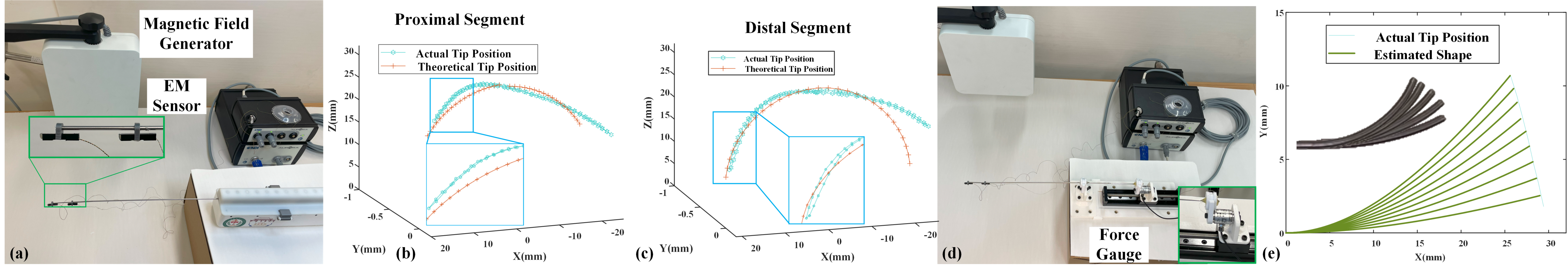}
	\caption{(a) Experimental setup for testing the mapping between push/pull distance and bending. (b)/(c) Tip position comparison between the actual value and the estimated ones via \eqref{Tip Position} for the proximal segment and the distal segment. (d) Experimental setup for testing the statics model. (e) Tip position comparison between the actual value and the estimated ones via \eqref{deflection} for the proximal segment.  }
	\label{fig:singlesegexpsetup}
\end{figure*}

\subsection{Single Segment Test}
Apart from hardware design, the basic performance of each segment should also be evaluated before putting into application. Herein, the rotation DoF is deemed as accurate, and only the mapping between push/pull distance and tip position is tested. Then, the statics model about push/pull force and the tip position is validated.\\ 
1) \texttt{Push/Pull Distance to Bending}

The whole actuation unit is fixed on our experimental platform, as shown in Fig. \ref{fig:singlesegexpsetup} (a), and four electromagnetic (EM) sensors (Aurora, Northern Digital Inc, Canada) are attached at the steerable section of each segment, to collect the actual tip position. The magnetic field generator covers the task space. Initially, the segment is in rest status ($D_p=0$), and then $D_p$ was commanded to increase to $2mm$ with an increment of $0.1mm$. In each increment, the tip position was collected 10 times and their mean value was used for validation. After reaching the maximum bending angle ($D_p=2mm$), it was decreased to $D_p=-2mm$, which means the steerable section is also in maximum bending status. At last, $D_p$ gradually reached the rest status. In summary, the variation of $D_p$ is: ${D_p} = 0 \to  2 \to  - 2 \to {D_p} = 0$, with the increment or decrement of 0.1mm. The theoretical tip position is obtained via \eqref{Tip Position}. For both proximal segment and the distal segment, the setting to the push/pull distance is identical. Fig. \ref{fig:singlesegexpsetup} (b) and (c) shows the comparison between the theoretical tip position and the actual values (collected by EM sensor) for the proximal segment and the distal segment, respectively. As a result, the errors between the estimated position and the actual are around 2.3mm and 2.7mm for the proximal segment and the distal segment, respectively. The larger error in the distal segment is due to the passive compliant segment, which is not exactly rigid and the push/pull force also acts on it. In addition, the EM sensor has a length of around 5mm, which also leads to measurement error. Each segment has a length of 30mm, and the tip position error to body length is around 7\% in open-loop manner, which is excellent compared with most existing slender robot arms.\\
2) \texttt{Axial Force to Bending}

To validate the statics model, we have also built another experimental platform to collect the actual data of actuation force and the tip position. As shown in Fig. \ref{fig:singlesegexpsetup} (d), the outer tube is fixed on the platform and the inner tube is fixed on a linear stage. The slider could be accurately controlled to generate translation, and further push or pull the inner tube. A single-axis force gauge (SBT641-19.6N, SIMBATUOCH, China) is connected with the inner tube. Similarly, the actual tip position is also collected by the EM sensing system. The inner tube is also pushed/pulled with the same actuation dataset, but in this experiment the actuation index is the axial force. Fig. \ref{fig:singlesegexpsetup} (e) shows the comparison  between the actual tip position and the estimated for the proximal segment. For brevity, the results of the distal segment are not presented. In addition to the tip position, the deflection curve is also almost consistent with the actual backbone. This further demonstrates the feasibility and accuracy of the proposed statics model.\\
\begin{figure}
	\centering
	\includegraphics[width=1\linewidth]{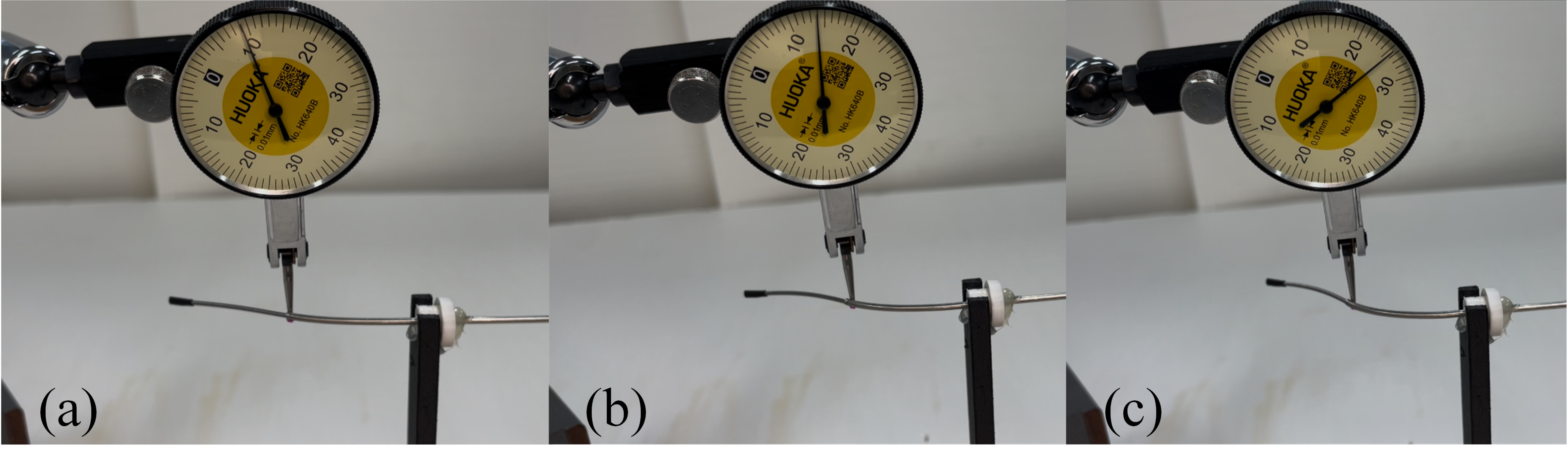}
	\caption{Experimental results of repetitive motion accuracy for the proximal segment. (a) $D_p^p=0.5mm$, (b) $D_p^p=1.5mm$, (c) $D_p^p=2mm$.}
	\label{fig:repetitiveresults}
\end{figure}
3) \texttt{Repetitive Motion Accuracy Test}

Validating the mapping between push/pull distance to bending angle is basic for the whole kinematics. Then, we also tested its repetitive motion accuracy, which is not only related to the tube itself but also related to the actuation unit. As shown in Fig. \ref{fig:repetitiveresults}, a dialgage (resolution: 0.01mm, 604B, Huoka, China) and actuation unit were fixed on a platform, ensuring that the tip of the proximal segment could touch the measurement area of the dialgage. The rigid section is also fixed to avoid waggle, and only the steerable section was commanded to bend. We set three different push distances, i.e., 0.5mm, 1.5mm and 2mm. In each condition, the actuation configuration was commanded three times to test the repetitive motion accuracy. Fig. \ref{fig:repetitiveresults} and supplementary video show the results. Consequently, the error are all within 0.01mm, which demonstrates the well designed actuation unit. In addition, with the increase of $ D_p^p$, the accuracy also increased, because larger bending angle promotes higher stiffness to against load.

\begin{figure}
	\centering
	\includegraphics[width=1\linewidth]{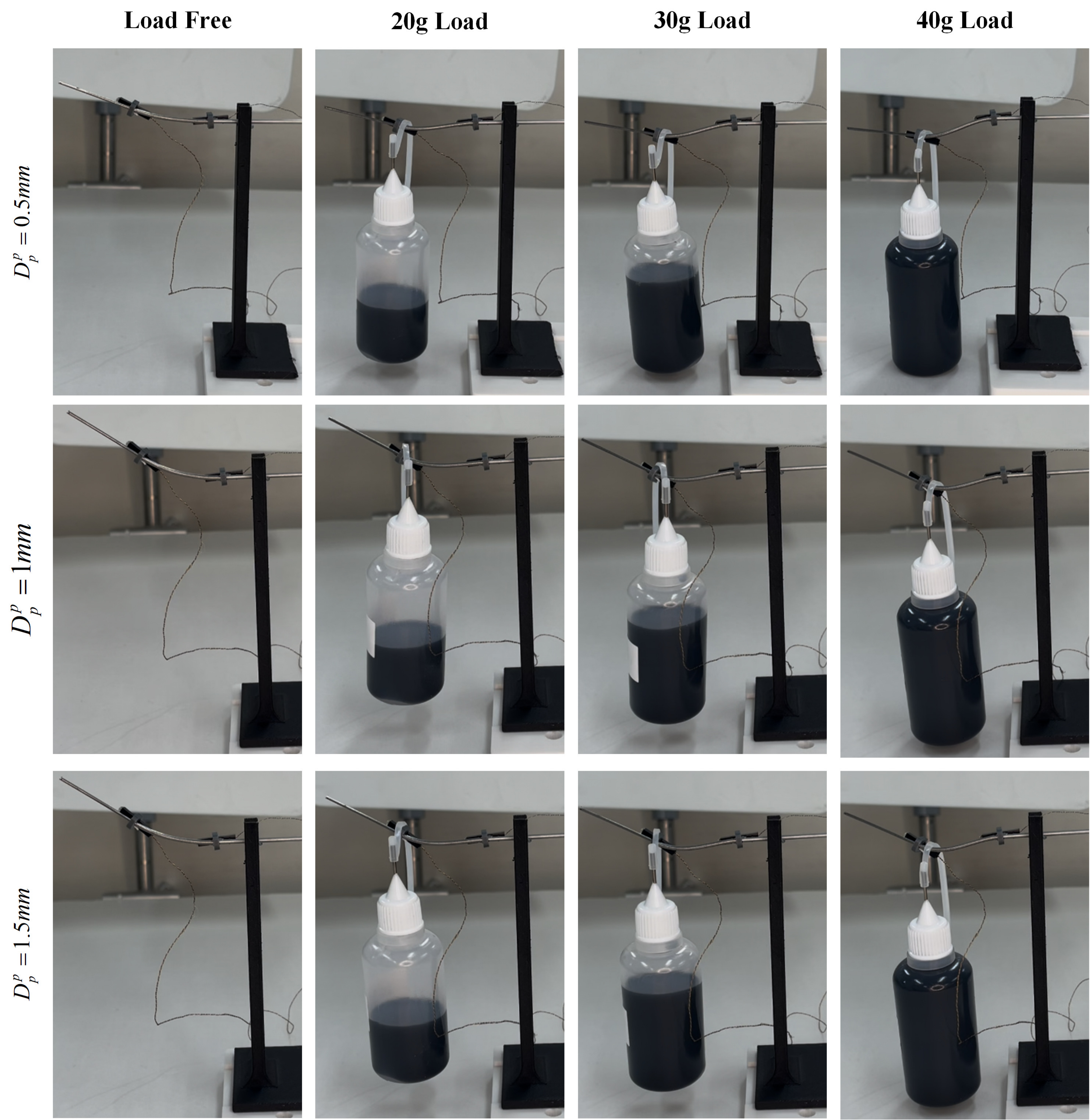}
	\caption{Load bearing test, with different actuation inputs and loads.}
	\label{fig:loadbearing}
\end{figure}
4) \texttt{Load-bearing Performance}

The stiffness of the steerable section is also an important index in actual application. We then tested the load-bearing performance of the proximal segment, because the distal segment has a long passive segment. Similarly, the outer tube is fixed and the inner tube is pushed to generate bending. The push distance $D_p^p$ was set to $ 0.5, 1$ and $1.5$ (unit: mm), and we collected the shape in load-free and with-load conditions. Tip position in the two scenarios were also collected for further detailed comparison. The tip position in load-free condition under a specific actuation push distance was used as standard to characterize the deviation with different loads. Fig. \ref{fig:loadbearing} shows the actual shape of the proximal segment in load-free and load conditions. Similarly, we fixed the rigid section and the actuation, with EM sensors collecting the tip position. Three plastic bottles filled with ink act as different loads to hang at the tip area. Then, the deviation between the tip position in load-free condition and load condition was calculated:
\begin{equation}
	{\rm{Dev}} (D_p^p)= \left\| {{P_{free}} - {P_{load}}} \right\|
\end{equation}
where ${P_{free}}$ and ${P_{load}}$ are respectively the tip position in load-free and with-load scenarios for a given actuation configuration $D_p^p$. TAB. \ref{tab:Dev} lists the deviation values for each setting.

\begin{table}[]
	\caption{Tip position deviation with comparison of load-free conditions.}
	\centering
	\label{tab:Dev}
	\begin{tabular}{p{1.85cm}<{\raggedright} p{0.25cm}<{\centering} p{0.3cm}<{\centering} p{0.35cm}<{\centering} p{0.3cm}<{\centering} p{0.3cm}<{\centering} p{0.35cm}<{\centering} p{0.25cm}<{\centering} p{0.25cm}<{\centering} p{0.3cm}<{\centering}}
		\hline
		$D_p^p$ (mm)        & \multicolumn{3}{c}{0.5} & \multicolumn{3}{c}{1} & \multicolumn{3}{c}{1.5} \\ \cline{2-10} 
		Load (g)       & 20     & 30     & 40    & 20    & 30    & 40    & 20     & 30     & 40    \\
		Deviation (mm) & 6.65   & 10.74   & 14.14  &7.05  & 10.52  & 11.83  & 6.22   & 8.83   & 11.99  \\ \hline
	\end{tabular}
\end{table}
 \begin{figure}
	\centering
	\includegraphics[width=1\linewidth]{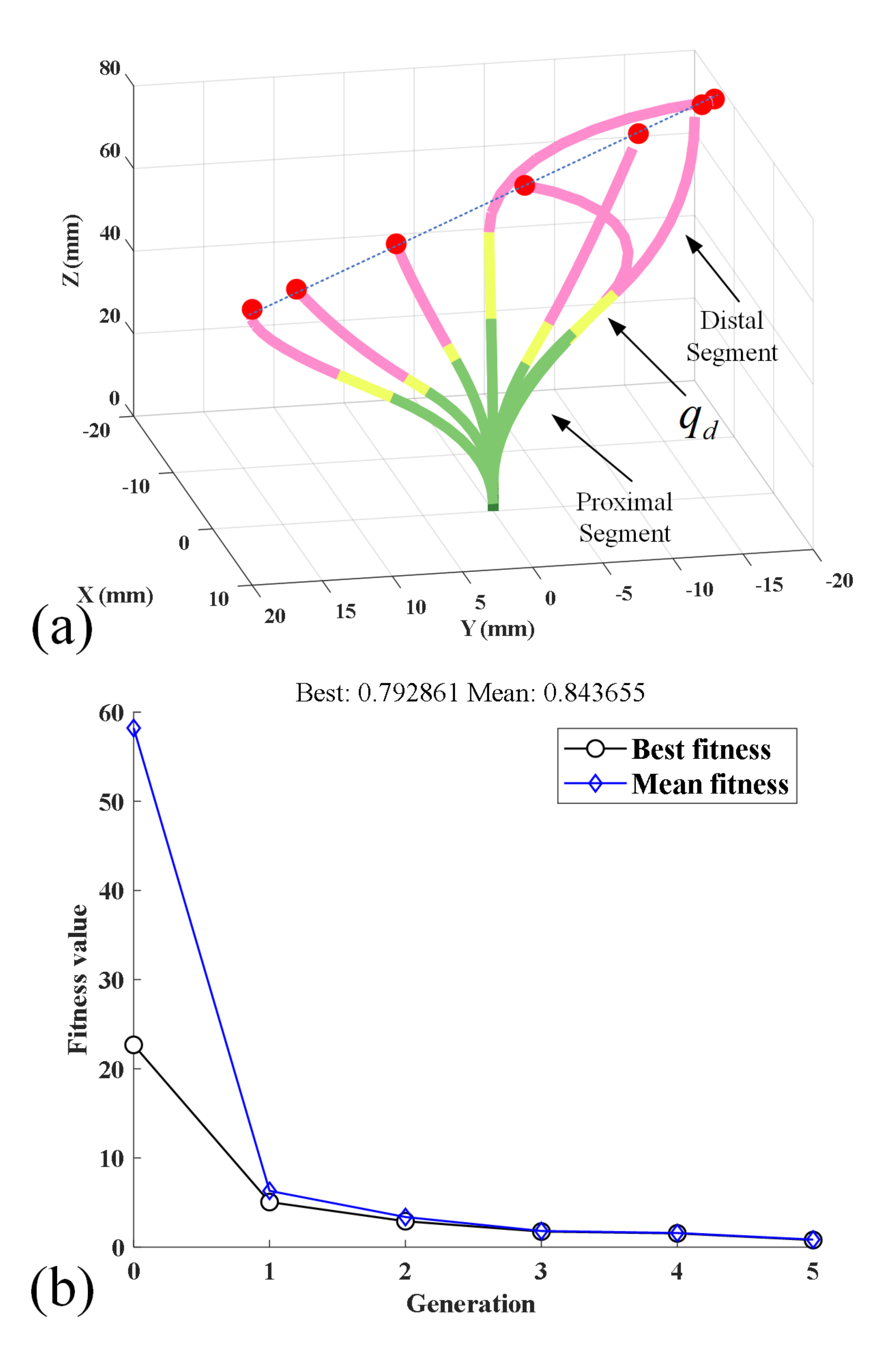}
	\caption{Inverse kinematics testing results. (a) Shape and tip position for a series of line path points. (b) Iteration process of GA to find the actuation inputs.}
	\label{fig:ikres}
\end{figure}
With the increase of load, the deviation increased accordingly. While robot in larger bending angle presents better load-against capability. Especially, when the $D_p^p$ was only 0.5mm, the robot arm could not compensate for the load with 40g (In Fig. \ref{fig:loadbearing}, 1st row and 4th column, the load bottle touched ground). This also demonstrate that larger bending angle has higher stiffness in manipulation. Load-bearing test on the distal segment was neglected for brevity and more importantly for the distal segment has a compliant section to bring the surgical energy (like laser or endoscopic camera) to a target area.

\subsection{Inverse Kinematics Validation}
Both the geometry-based and statics models have been tested, and the results show the accuracy of the proposed approaches. Then, the distal segment passes through the hollow space of the proximal segment, forming the dual-segment design. Considering the hyper redundancy of the kinematics mapping, we focused on testing the inverse kinematics model, i.e., accuracy of finding the optimal actuation inputs for a given desired tip position. As shown in Fig. \ref{fig:ikres} (a), six path points were given to form a line path, which are given desired tip position. Then, the system worked to find the optimal actuation inputs. In Fig. \ref{fig:ikres} (a), each shape is well mapped with the task points, and the whole process only takes 0.09s (MATLAB 2022a) for searching one instance. In term of parameters in GA, the size of population was set to 20, and elite number was 10 in searching. As shown in Fig. \ref{fig:ikres} (b), the loss function almost converged after five iterations, so we set the maximum iteration step to 20 to ensure convergence.

\begin{figure*}[t]
	\centering
	\includegraphics[width=1\linewidth]{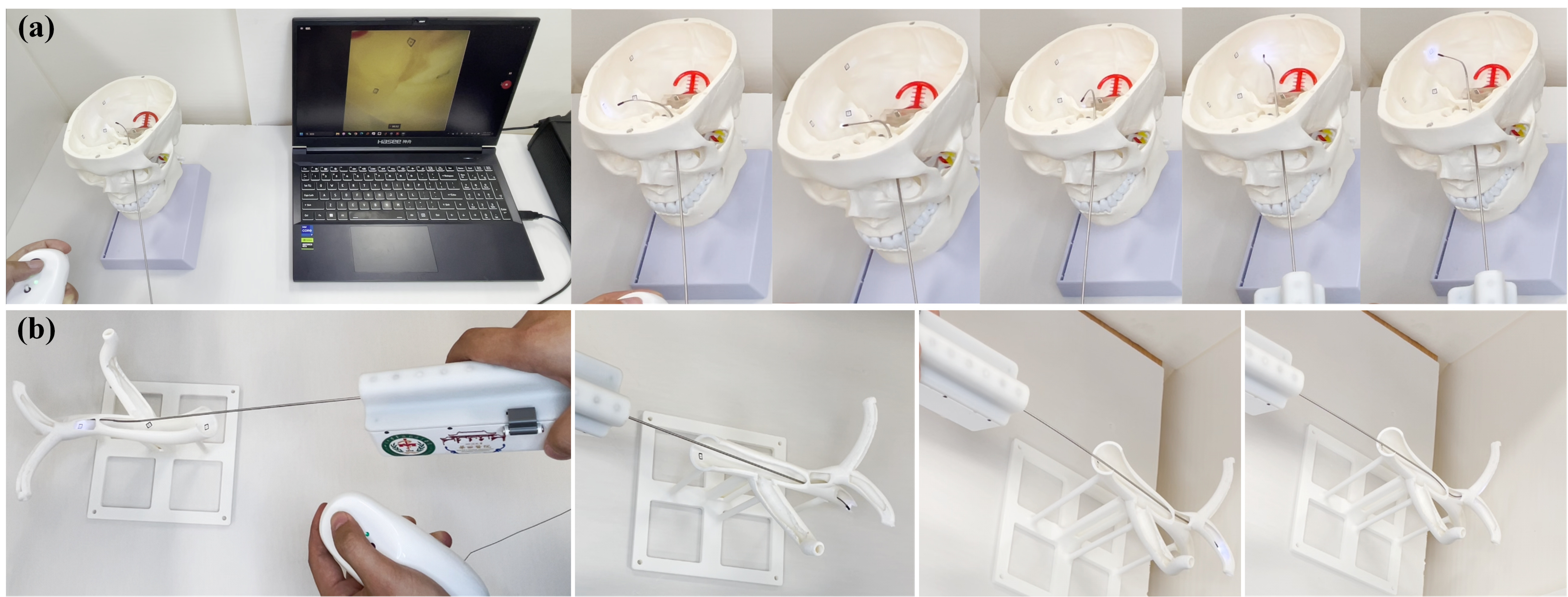}
	\caption{Maneuverability test. The dexterous robot has an endoscopic camera at the tip to explore narrow space. (a) Skull phantom to test trans-orbital surgery. (b) Explore a bronchus phantom. }
	\label{fig:maneuverability}
\end{figure*}

\subsection{Maneuverability Test}

While the two segments simultaneously reached the maximum bending angles, the end effector has large workspace and dexterity. An endoscopic camera (TA10, Omnivision) was fixed at the tip of the distal segment, which collects the view of task space. We then prepared two phantom experiments to test its overall maneuverability. Fig. \ref{fig:maneuverability} (a) shows the overall experimental setup. Operator holds the surgical device and the wireless joystick by his hands, and the laptop shows the camera view. In the skull, we tagged digits as landmarks, and the goal is to maneuver the robot arm to clearly see the digits. As can be seen in Fig. \ref{fig:maneuverability} (a), the robot arm could dexterously pass through the whole behind orbit, which is just around 3mm in diameter. Through tuning the robot shape, each landmark was gradually seen. The robot arm presented various curved shapes, which further demonstrates its dexterity.

In addition, considering the existing complexity in endoscopic bronchus examination, we attempted to leverage the dexterity to explore the zigzag path with collecting the view inside it. Similarly, a 3D-printed bronchial model was fixed on platform, and operator (second author) held the surgical device using his right hand. His left hand operated the wireless joystick to position the tip and the whole shape, which simulates the bronchscopic health examination. There are totally three zigzag curved paths to explore. Initially, the flexible arm entered from the main trunk, and the translation motion of the whole device is manipulated by operator. When it moved beside the intersection, the distal segment was first commanded to bend towards the desired branch. As shown in Fig. \ref{fig:maneuverability} (b), the straight proximal segment was then controlled to present a curved shape to adapt with the curved path. Consequently, the flexible robot could smoothly enter the three branches. The whole process was recorded in the supplementary video.

\subsection{Potential Application in Minimally Invasive Surgery}
\begin{figure}
	\centering
	\includegraphics[width=1\linewidth]{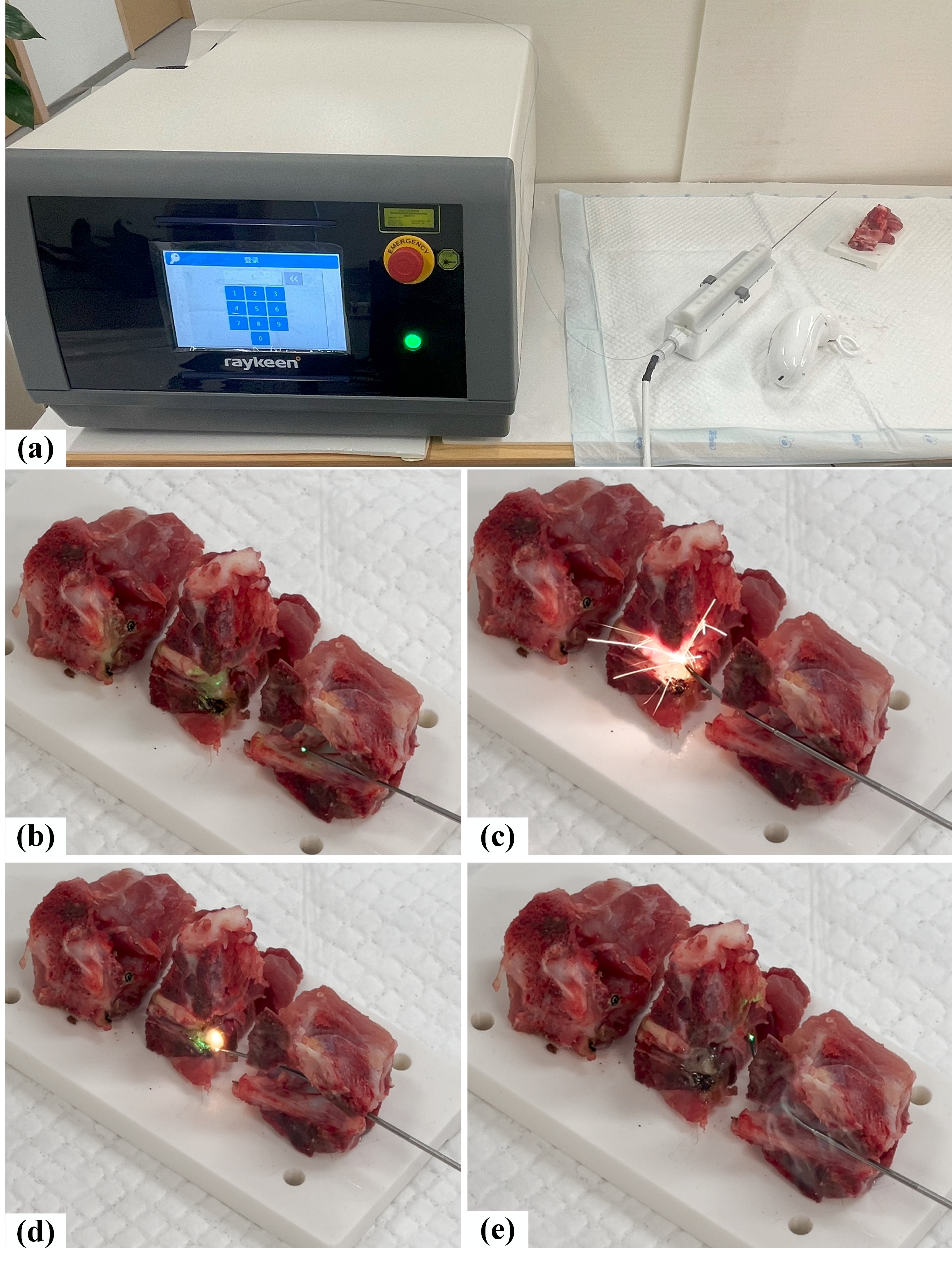}
	\caption{Potential in Minimally Invasive Surgery. (a) Experimental setup. (b)-(e) Process of cauterizing bone.}
	\label{fig:misapplication}
\end{figure}

Beside the endoscopic camera, surgical manipulation is also generally indispensable in applications. One of the notable advantages of the flexible robot arm is its ultra-slender robot body with comparable inner hollow space. Fig. \ref{fig:misapplication} shows the experimental setup. A thulium laser device (Raykeen,  China) generates energy to perform bone decompressing, and a laser fiber (0.4mm in diameter) passes through the hollow lumen of the robot arm. Three pork spines are arrayed on the platform, and the robot was also handheld. Another operator was responsible for delivering the laser fiber. The robot was manipulated to pass the narrow gap of the spine bone. As shown in Fig. \ref{fig:misapplication} (b), the laser fiber was guided by the distal segment. In Fig. \ref{fig:misapplication} (c) and (d), the laser was activated to perform cauterization. Finally, the laser stopped and robot retracted from the passage.

\section{Conclusion}\label{Conclusion}
In this work, we proposed a novel co-axial antagonistic concentric tubular robot on the basis of concentric push/pull robot arm. Delicately designed tenon-mortise slits were proposed to enhance its overall stiffness, and corresponding theoretical design scheme considering dexterity and stiffness was presented. In addition, we analyzed the mappings of push/pull distance-bending angle and axial actuation force-bending angle, forming the kinetostatic model for the robot arm. For the dual-segment configuration with more dexterity, we proposed kinematics model based on constrained optimization algorithm to bring the robot tip to a given desired task configuration. In terms of the robot performance, we analyzed its task space and proposed a dexterity evaluation method. With a novel compact actuation unit, the six-DoF slender robot arm was well actuated. Based on that, we conducted various experiments to evaluate its performance, including validating the models, basic motion accuracy tests, stiffness test, maneuverability and more importantly demonstrating its application in Minimally Invasive Surgeries.

To the best of our knowledge, this is the first small-sized slender robot arm with notable stiffness large hollow space ratio. The proposed design scheme, control model and actuation unit could also be applied in similar continuum robotics. In the future, we will focus on investigating the proprioceptive mechanism to collect the real-time shape in manipulation.

\section{Acknowledge}
We would like to express our sincere gratitude to the Medical Simulation Center of West China Hospital for providing the testing facilities that supported this study.

\bibliographystyle{IEEEtran}
\bibliography{REF}

\begin{thebibliography}{10}
\providecommand{\url}[1]{#1}
\csname url@samestyle\endcsname
\providecommand{\newblock}{\relax}
\providecommand{\bibinfo}[2]{#2}
\providecommand{\BIBentrySTDinterwordspacing}{\spaceskip=0pt\relax}
\providecommand{\BIBentryALTinterwordstretchfactor}{4}
\providecommand{\BIBentryALTinterwordspacing}{\spaceskip=\fontdimen2\font plus
\BIBentryALTinterwordstretchfactor\fontdimen3\font minus
  \fontdimen4\font\relax}
\providecommand{\BIBforeignlanguage}[2]{{%
\expandafter\ifx\csname l@#1\endcsname\relax
\typeout{** WARNING: IEEEtran.bst: No hyphenation pattern has been}%
\typeout{** loaded for the language `#1'. Using the pattern for}%
\typeout{** the default language instead.}%
\else
\language=\csname l@#1\endcsname
\fi
#2}}
\providecommand{\BIBdecl}{\relax}
\BIBdecl

\bibitem{wang2023novel}
J.~Wang, C.~Hu, G.~Ning, L.~Ma, X.~Zhang, and H.~Liao, ``A novel miniature
  spring-based continuum manipulator for minimally invasive surgery: Design and
  evaluation,'' \emph{IEEE/ASME Transactions on Mechatronics}, vol.~28, no.~5,
  pp. 2716--2727, 2023.

\bibitem{dupont2022continuum}
P.~E. Dupont, N.~Simaan, H.~Choset, and C.~Rucker, ``Continuum robots for
  medical interventions,'' \emph{Proceedings of the IEEE}, vol. 110, no.~7, pp.
  847--870, 2022.

\bibitem{russo2023continuum}
M.~Russo, ``Continuum robots for space applications,'' in \emph{Design Advances
  in Aerospace Robotics: Proceedings of TORVEASTRO 2023}.\hskip 1em plus 0.5em
  minus 0.4em\relax Springer, 2023, pp. 129--139.

\bibitem{zhang2022novel}
J.~Zhang, Z.~Kan, Y.~Li, Z.~Wu, J.~Wu, and H.~Peng, ``Novel design of a
  cable-driven continuum robot with multiple motion patterns,'' \emph{IEEE
  Robotics and Automation Letters}, vol.~7, no.~3, pp. 6163--6170, 2022.

\bibitem{jalali2024dynamic}
A.~Jalali and F.~Janabi-Sharifi, ``Dynamic manipulation and stiffness
  modulation of cooperative continuum robots: Theory and experiment,''
  \emph{Journal of Mechanisms and Robotics}, vol.~16, no.~12, p. 121001, 2024.

\bibitem{zhang2022integrated}
P.~Zhang, I.~M. Lei, G.~Chen, J.~Lin, X.~Chen, J.~Zhang, C.~Cai, X.~Liang, and
  J.~Liu, ``Integrated 3d printing of flexible electroluminescent devices and
  soft robots,'' \emph{Nature Communications}, vol.~13, no.~1, p. 4775, 2022.

\bibitem{shen2023design}
D.~Shen, Q.~Zhang, Y.~Han, C.~Tu, and X.~Wang, ``Design and development of a
  continuum robot with switching-stiffness,'' \emph{Soft Robotics}, vol.~10,
  no.~5, pp. 1015--1027, 2023.

\bibitem{tummers2023cosserat}
M.~Tummers, V.~Lebastard, F.~Boyer, J.~Troccaz, B.~Rosa, and M.~T. Chikhaoui,
  ``Cosserat rod modeling of continuum robots from newtonian and lagrangian
  perspectives,'' \emph{IEEE Transactions on Robotics}, 2023.

\bibitem{lilge2022kinetostatic}
S.~Lilge and J.~Burgner-Kahrs, ``Kinetostatic modeling of tendon-driven
  parallel continuum robots,'' \emph{IEEE Transactions on Robotics}, 2022.

\bibitem{zhong2020recent}
Y.~Zhong, L.~Hu, and Y.~Xu, ``Recent advances in design and actuation of
  continuum robots for medical applications,'' in \emph{Actuators}, vol.~9,
  no.~4.\hskip 1em plus 0.5em minus 0.4em\relax MDPI, 2020, p. 142.

\bibitem{zhao2021reconstructing}
Q.~Zhao, J.~Lai, and H.~K. Chu, ``Reconstructing external force on the
  circumferential body of continuum robot with embedded proprioceptive
  sensors,'' \emph{IEEE Transactions on Industrial Electronics}, vol.~69,
  no.~12, pp. 13\,111--13\,120, 2021.

\bibitem{kim2019ferromagnetic}
Y.~Kim, G.~A. Parada, S.~Liu, and X.~Zhao, ``Ferromagnetic soft continuum
  robots,'' \emph{Science Robotics}, vol.~4, no.~33, p. eaax7329, 2019.

\bibitem{kato2014tendon}
T.~Kato, I.~Okumura, S.-E. Song, A.~J. Golby, and N.~Hata, ``Tendon-driven
  continuum robot for endoscopic surgery: Preclinical development and
  validation of a tension propagation model,'' \emph{IEEE/ASME Transactions on
  Mechatronics}, vol.~20, no.~5, pp. 2252--2263, 2014.

\bibitem{chitalia2020towards}
Y.~Chitalia, N.~J. Deaton, S.~Jeong, N.~Rahman, and J.~P. Desai, ``Towards
  fbg-based shape sensing for micro-scale and meso-scale continuum robots with
  large deflection,'' \emph{IEEE robotics and automation letters}, vol.~5,
  no.~2, pp. 1712--1719, 2020.

\bibitem{chitalia2018design}
Y.~Chitalia, X.~Wang, and J.~P. Desai, ``Design, modeling and control of a
  2-dof robotic guidewire,'' in \emph{2018 IEEE International Conference on
  Robotics and Automation (ICRA)}.\hskip 1em plus 0.5em minus 0.4em\relax IEEE,
  2018, pp. 32--37.

\bibitem{mc2020continuum}
C.~Mc~Caffrey, T.~Umedachi, W.~Jiang, T.~Sasatani, Y.~Narusue, R.~Niiyama, and
  Y.~Kawahara, ``Continuum robotic caterpillar with wirelessly powered shape
  memory alloy actuators,'' \emph{Soft robotics}, vol.~7, no.~6, pp. 700--710,
  2020.

\bibitem{campisano2021closed}
F.~Campisano, S.~Cal{\'o}, A.~A. Remirez, J.~H. Chandler, K.~L. Obstein, R.~J.
  Webster~III, and P.~Valdastri, ``Closed-loop control of soft continuum
  manipulators under tip follower actuation,'' \emph{The International journal
  of robotics research}, vol.~40, no. 6-7, pp. 923--938, 2021.

\bibitem{tutcu2021quasi}
C.~Tutcu, B.~A. Baydere, S.~K. Talas, and E.~Samur, ``Quasi-static modeling of
  a novel growing soft-continuum robot,'' \emph{The International Journal of
  Robotics Research}, vol.~40, no.~1, pp. 86--98, 2021.

\bibitem{nwafor2023design}
C.~J. Nwafor, C.~Girerd, G.~J. Laurent, T.~K. Morimoto, and K.~Rabenorosoa,
  ``Design and fabrication of concentric tube robots: A survey,'' \emph{IEEE
  Transactions on Robotics}, 2023.

\bibitem{stilli2014shrinkable}
A.~Stilli, H.~A. Wurdemann, and K.~Althoefer, ``Shrinkable,
  stiffness-controllable soft manipulator based on a bio-inspired antagonistic
  actuation principle,'' in \emph{2014 IEEE/RSJ International Conference on
  Intelligent Robots and Systems}.\hskip 1em plus 0.5em minus 0.4em\relax IEEE,
  2014, pp. 2476--2481.

\bibitem{yang2020geometric}
C.~Yang, S.~Geng, I.~Walker, D.~T. Branson, J.~Liu, J.~S. Dai, and R.~Kang,
  ``Geometric constraint-based modeling and analysis of a novel continuum robot
  with shape memory alloy initiated variable stiffness,'' \emph{The
  International Journal of Robotics Research}, vol.~39, no.~14, pp. 1620--1634,
  2020.

\bibitem{pang2020coboskin}
G.~Pang, G.~Yang, W.~Heng, Z.~Ye, X.~Huang, H.-Y. Yang, and Z.~Pang,
  ``Coboskin: Soft robot skin with variable stiffness for safer human--robot
  collaboration,'' \emph{IEEE Transactions on Industrial Electronics}, vol.~68,
  no.~4, pp. 3303--3314, 2020.

\bibitem{langer2018stiffening}
M.~Langer, E.~Amanov, and J.~Burgner-Kahrs, ``Stiffening sheaths for continuum
  robots,'' \emph{Soft robotics}, vol.~5, no.~3, pp. 291--303, 2018.

\bibitem{herzig2018variable}
N.~Herzig, P.~Maiolino, F.~Iida, and T.~Nanayakkara, ``A variable stiffness
  robotic probe for soft tissue palpation,'' \emph{IEEE Robotics and Automation
  Letters}, vol.~3, no.~2, pp. 1168--1175, 2018.

\bibitem{zhang2023bioinspired}
J.~Zhang, B.~Wang, H.~Chen, J.~Bai, Z.~Wu, J.~Liu, H.~Peng, and J.~Wu,
  ``Bioinspired continuum robots with programmable stiffness by harnessing
  phase change materials,'' \emph{Advanced Materials Technologies}, vol.~8,
  no.~6, p. 2201616, 2023.

\bibitem{wang2019variable}
H.~Wang, Z.~Du, W.~Yang, Z.~Y. Yan, and X.~Wang, ``Variable stiffness model
  construction and simulation verification of coupled notch continuum
  manipulator,'' \emph{IEEE Access}, vol.~7, pp. 154\,761--154\,769, 2019.

\bibitem{oliver2021concentric}
K.~Oliver-Butler, J.~A. Childs, A.~Daniel, and D.~C. Rucker, ``Concentric
  push--pull robots: Planar modeling and design,'' \emph{IEEE Transactions on
  Robotics}, vol.~38, no.~2, pp. 1186--1200, 2021.

\bibitem{oliver2017concentric}
K.~Oliver-Butler, Z.~H. Epps, and D.~C. Rucker, ``Concentric agonist-antagonist
  robots for minimally invasive surgeries,'' in \emph{Medical Imaging 2017:
  Image-Guided Procedures, Robotic Interventions, and Modeling}, vol.
  10135.\hskip 1em plus 0.5em minus 0.4em\relax SPIE, 2017, pp. 270--278.

\bibitem{park2021design}
S.~Park, J.~Kim, C.~Kim, K.-J. Cho, and G.~Noh, ``Design optimization of
  asymmetric patterns for variable stiffness of continuum tubular robots,''
  \emph{IEEE Transactions on Industrial Electronics}, vol.~69, no.~8, pp.
  8190--8200, 2021.

\bibitem{barrientos2023asymmetric}
J.~Barrientos-Diez, M.~Russo, X.~Dong, D.~Axinte, and J.~Kell, ``Asymmetric
  continuum robots,'' \emph{IEEE Robotics and Automation Letters}, vol.~8,
  no.~3, pp. 1279--1286, 2023.

\bibitem{wei2012modeling}
D.~Wei, Y.~Wenlong, H.~Dawei, and D.~Zhijiang, ``Modeling of flexible arm with
  triangular notches for applications in single port access abdominal
  surgery,'' in \emph{2012 IEEE International Conference on Robotics and
  Biomimetics (ROBIO)}.\hskip 1em plus 0.5em minus 0.4em\relax IEEE, 2012, pp.
  588--593.

\bibitem{bell2012deflectable}
J.~A. Bell, C.~E. Saikus, K.~Ratnayaka, V.~Wu, M.~Sonmez, A.~Z. Faranesh, J.~H.
  Colyer, R.~J. Lederman, and O.~Kocaturk, ``A deflectable guiding catheter for
  real-time mri-guided interventions,'' \emph{Journal of Magnetic Resonance
  Imaging}, vol.~35, no.~4, pp. 908--915, 2012.

\bibitem{kim2019continuously}
J.~Kim, W.-Y. Choi, S.~Kang, C.~Kim, and K.-J. Cho, ``Continuously variable
  stiffness mechanism using nonuniform patterns on coaxial tubes for continuum
  microsurgical robot,'' \emph{IEEE Transactions on Robotics}, vol.~35, no.~6,
  pp. 1475--1487, 2019.

\bibitem{webster2010design}
R.~J. Webster~III and B.~A. Jones, ``Design and kinematic modeling of constant
  curvature continuum robots: A review,'' \emph{The International Journal of
  Robotics Research}, vol.~29, no.~13, pp. 1661--1683, 2010.

\bibitem{he2018research}
B.~He, S.~Xu, and Z.~Wang, ``Research on stiffness of multibackbone continuum
  robot based on screw theory and euler-bernoulli beam,'' \emph{Mathematical
  Problems in Engineering}, vol. 2018, 2018.

\bibitem{singh2021dynamic}
P.~K. Singh and K.~C. Murali, ``Dynamic characteristics of continuum robot for
  colonoscopy using finite element simulation,'' \emph{Proceedings of the
  Institution of Mechanical Engineers, Part C: Journal of Mechanical
  Engineering Science}, vol. 235, no.~22, pp. 6398--6414, 2021.

\bibitem{jensen2022tractable}
S.~W. Jensen, C.~C. Johnson, A.~M. Lindberg, and M.~D. Killpack, ``Tractable
  and intuitive dynamic model for soft robots via the recursive newton-euler
  algorithm,'' in \emph{2022 IEEE 5th International Conference on Soft Robotics
  (RoboSoft)}.\hskip 1em plus 0.5em minus 0.4em\relax IEEE, 2022, pp. 416--422.

\bibitem{falkenhahn2014dynamic}
V.~Falkenhahn, T.~Mahl, A.~Hildebrandt, R.~Neumann, and O.~Sawodny, ``Dynamic
  modeling of constant curvature continuum robots using the euler-lagrange
  formalism,'' in \emph{2014 IEEE/RSJ international conference on intelligent
  robots and systems}.\hskip 1em plus 0.5em minus 0.4em\relax IEEE, 2014, pp.
  2428--2433.

\bibitem{rone2013continuum}
W.~S. Rone and P.~Ben-Tzvi, ``Continuum robot dynamics utilizing the principle
  of virtual power,'' \emph{IEEE Transactions on Robotics}, vol.~30, no.~1, pp.
  275--287, 2013.

\bibitem{mattson2004smart}
C.~A. Mattson, A.~A. Mullur, and A.~Messac, ``Smart pareto filter: Obtaining a
  minimal representation of multiobjective design space,'' \emph{Engineering
  Optimization}, vol.~36, no.~6, pp. 721--740, 2004.

\end{thebibliography}

\vspace{-1.2cm}
\begin{IEEEbiography}[{\includegraphics[width=1in,height=1.25in,clip,keepaspectratio]{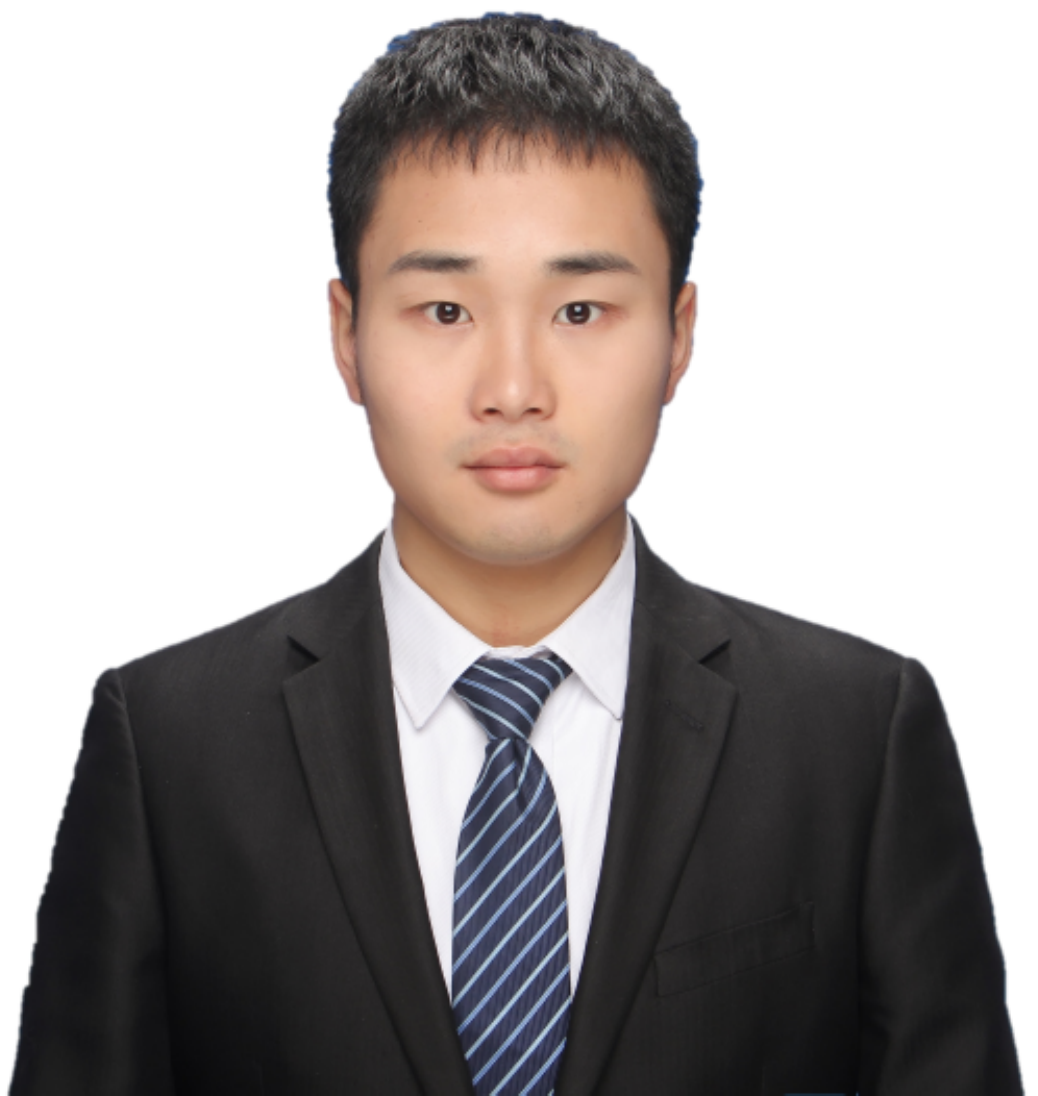}}]
	{Qingxiang Zhao} received the B.Eng. degree and the M.S. degree in mechanical engineering from Sichuan University, Chengdu, China, in 2016 and 2019, respectively. He obtained Ph.D. degree in mechanical engineering from the Hong Kong Polytechnic University in 2022. He was an assistant professor in the Centre for Artificial Intelligence and Robotics (CAIR) Hong Kong Institute of Science \& Innovation, Chinese Academy of Sciences. He is currently a postdoctoral fellow with the West China School of Medicine, Sichuan University.	His research interests include soft robotics and surgical robotics.
\end{IEEEbiography}
\vspace{-1.2cm}

\begin{IEEEbiography}[{\includegraphics[width=1in,height=1.25in,clip,keepaspectratio]{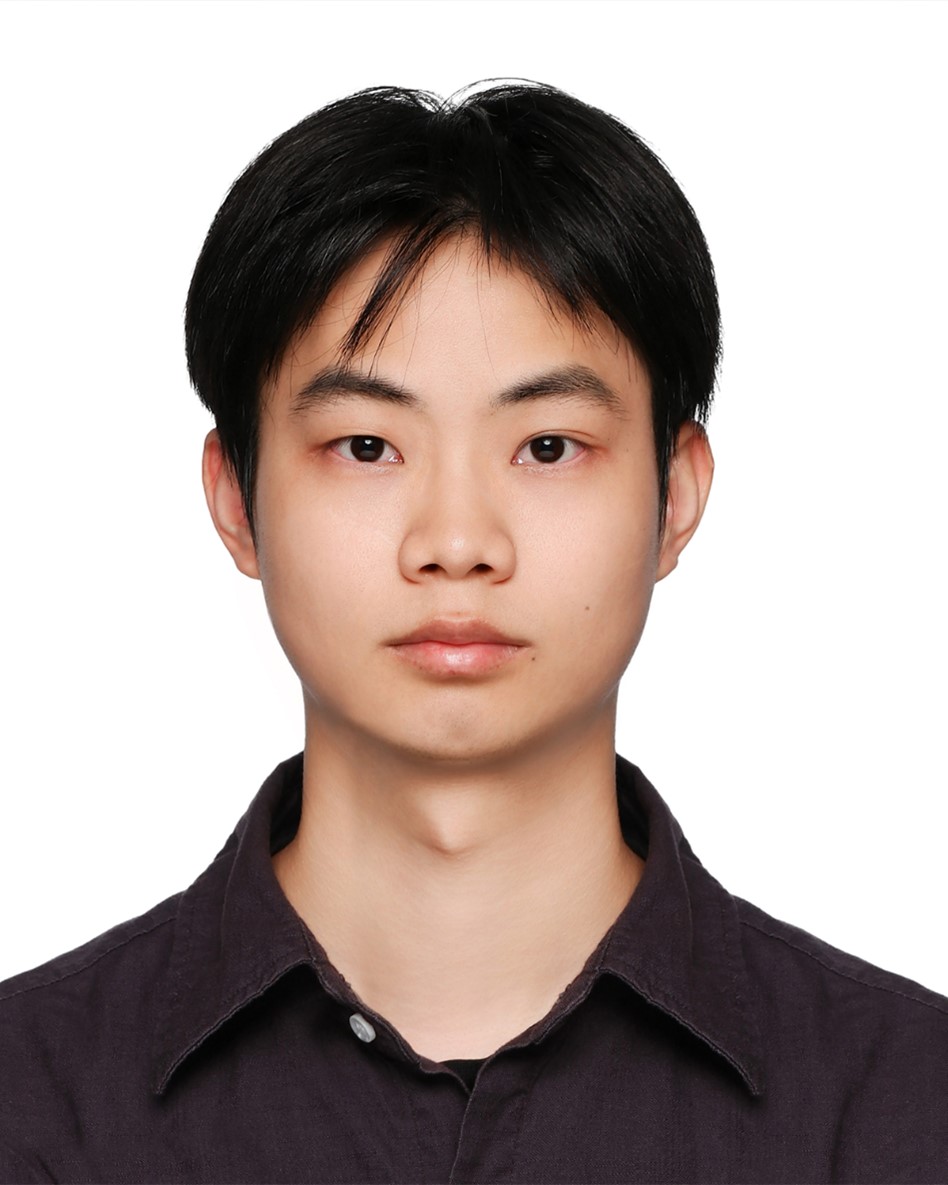}}]
	{Runfeng Zhu} received the B.Eng. degree in Industrial Engineering from the Guangdong University of Technology, Guangzhou, China in 2021 and the M.Sc. degree in Mechanical Engineering from the Hong Kong Polytechnic University, Hong Kong SAR China in 2024.
	Currently, he is working toward the Ph.D. degree in mechanical engineering in Shien-Ming Wu school of Intelligent Engineering with the South China University of Technology, Guangzhou, China. His research interests include soft robotics, surgical robots and Biorobotics.
\end{IEEEbiography}

\vspace{-1.2cm}
\begin{IEEEbiography}[{\includegraphics[width=1in,height=1.25in,clip,keepaspectratio]{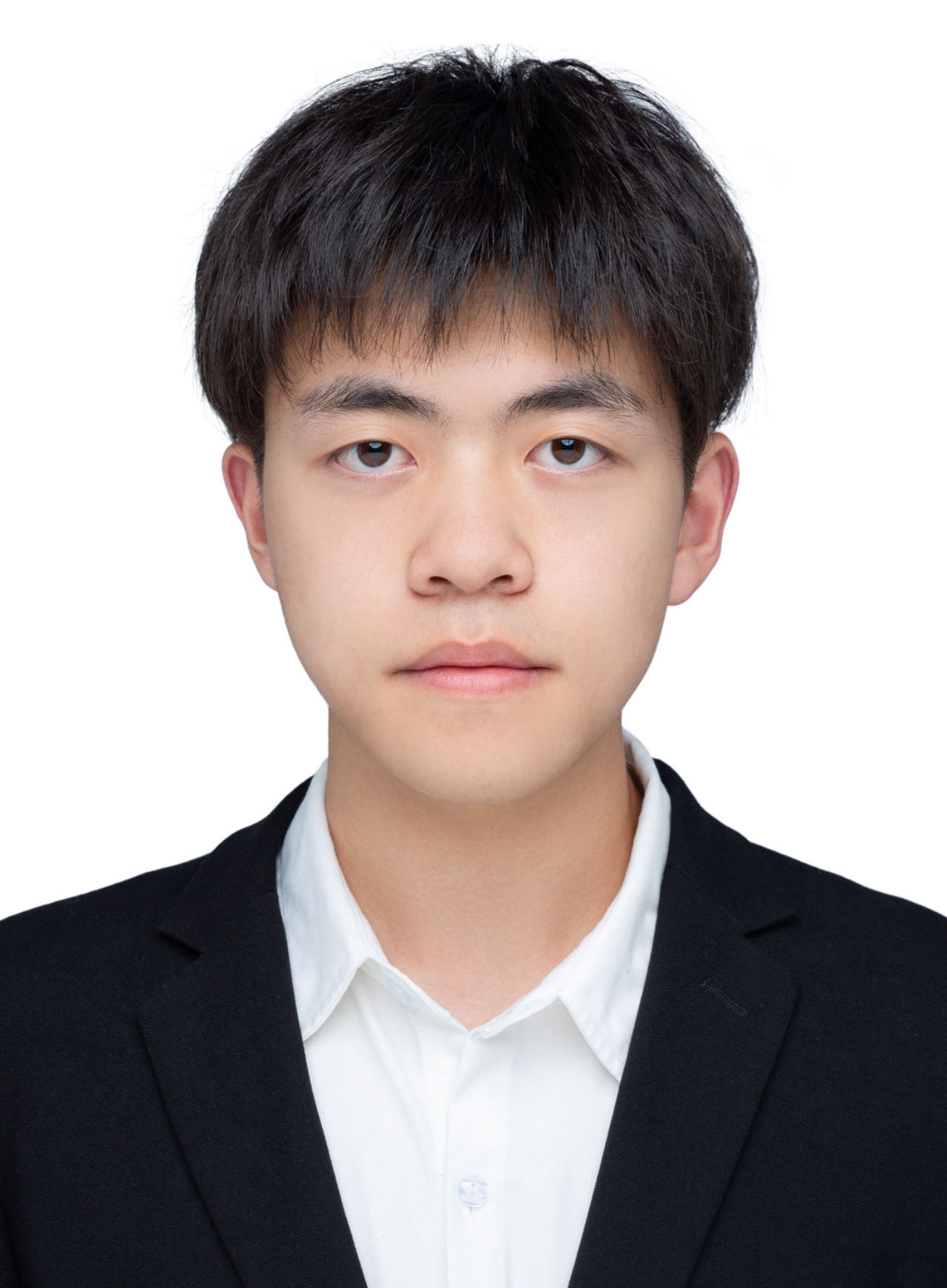}}]
	{Xin Zhong} is currently an undergraduate student majoring in Mechanical Engineering at School of Mechanical and Electrical Engineering, University of Electronic Science and Technology of China, Chengdu, China from 2022. His research interests include continuum robots, soft body robots and robot collaboration, as well as modeling and controlling strategies of embedded systems.
\end{IEEEbiography}
\vspace{-1.2cm}
\begin{IEEEbiography}[{\includegraphics[width=1in,height=1.25in,clip,keepaspectratio]{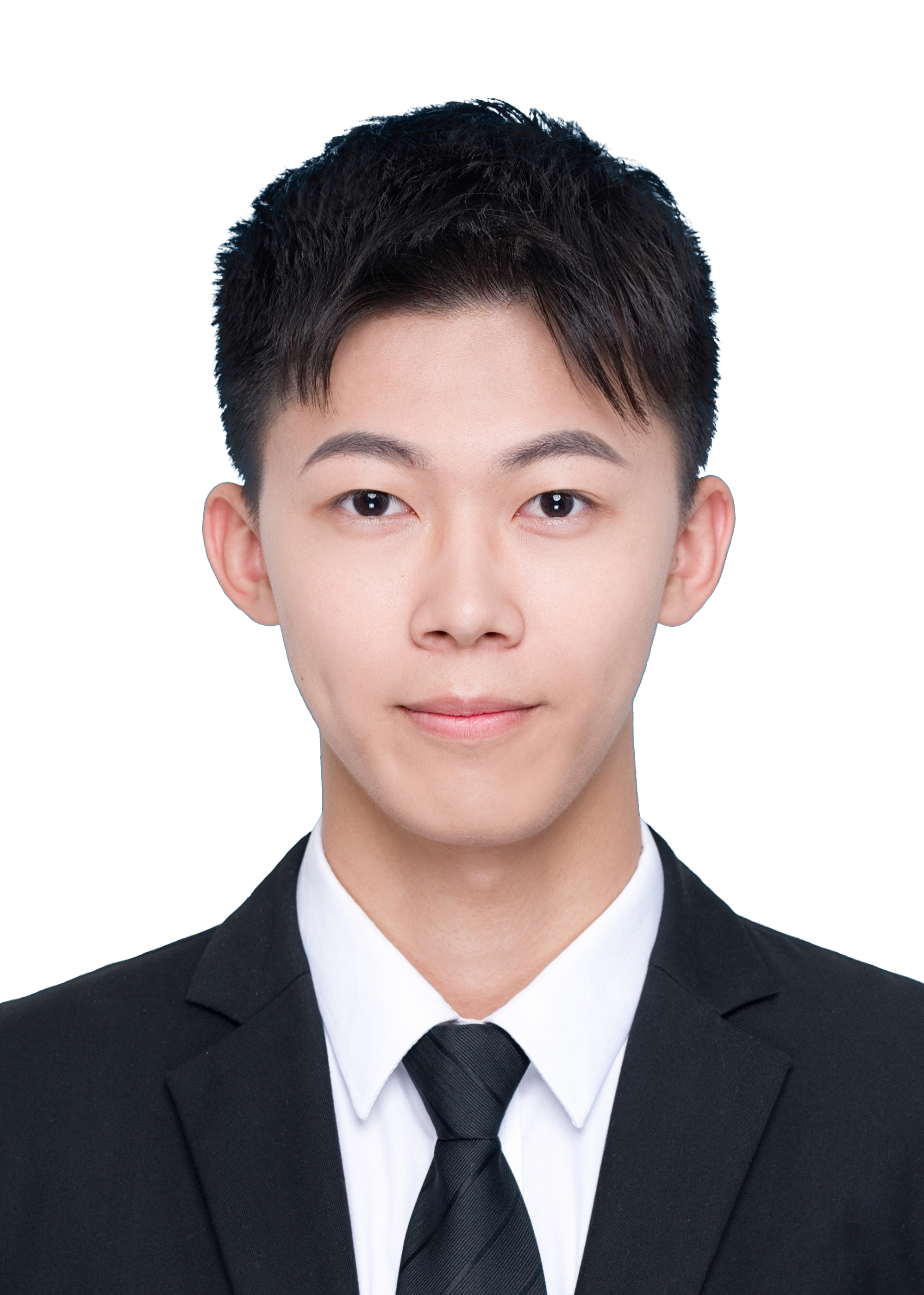}}]
	{Baitao Lin} received the B.Eng. degree in mechanical engineering from the Guangdong University of
	Technology, Guangzhou, China in 2023. He is currently in his second year of study for his Master's degree at School of Mechanical Engineering, Sichuan University, Chengdu, China. His research
	interests include soft robotics and surgical robotics.
\end{IEEEbiography}
\vspace{-1.2cm}
\begin{IEEEbiography}[{\includegraphics[width=1in,height=1.25in,clip,keepaspectratio]{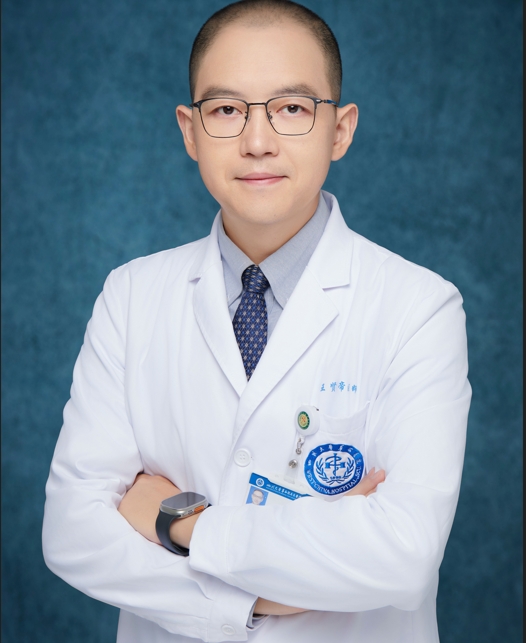}}] {Xiandi Wang} received a Ph.D. in Orthopedic Surgery from Fudan University, China, and an M.D. from Binzhou Medical University, China. He completed his residency in Orthopedic Surgery at West China Hospital (WCH), Sichuan University, China and further specialized as a spine surgery fellow at Fudan University. Currently, he serves as an attending surgeon at WCH and is also affiliated with its Medical Simulation Center. His research interests include spinal surgery, with a focus on minimally invasive techniques, medical education emphasizing simulation-based learning, and the integration of biomedical engineering, including medical robotics, surgical navigation, and artificial intelligence..
\end{IEEEbiography}

\vspace{-1.2cm}
\begin{IEEEbiography}[{\includegraphics[width=1in,height=1.25in, clip, keepaspectratio]{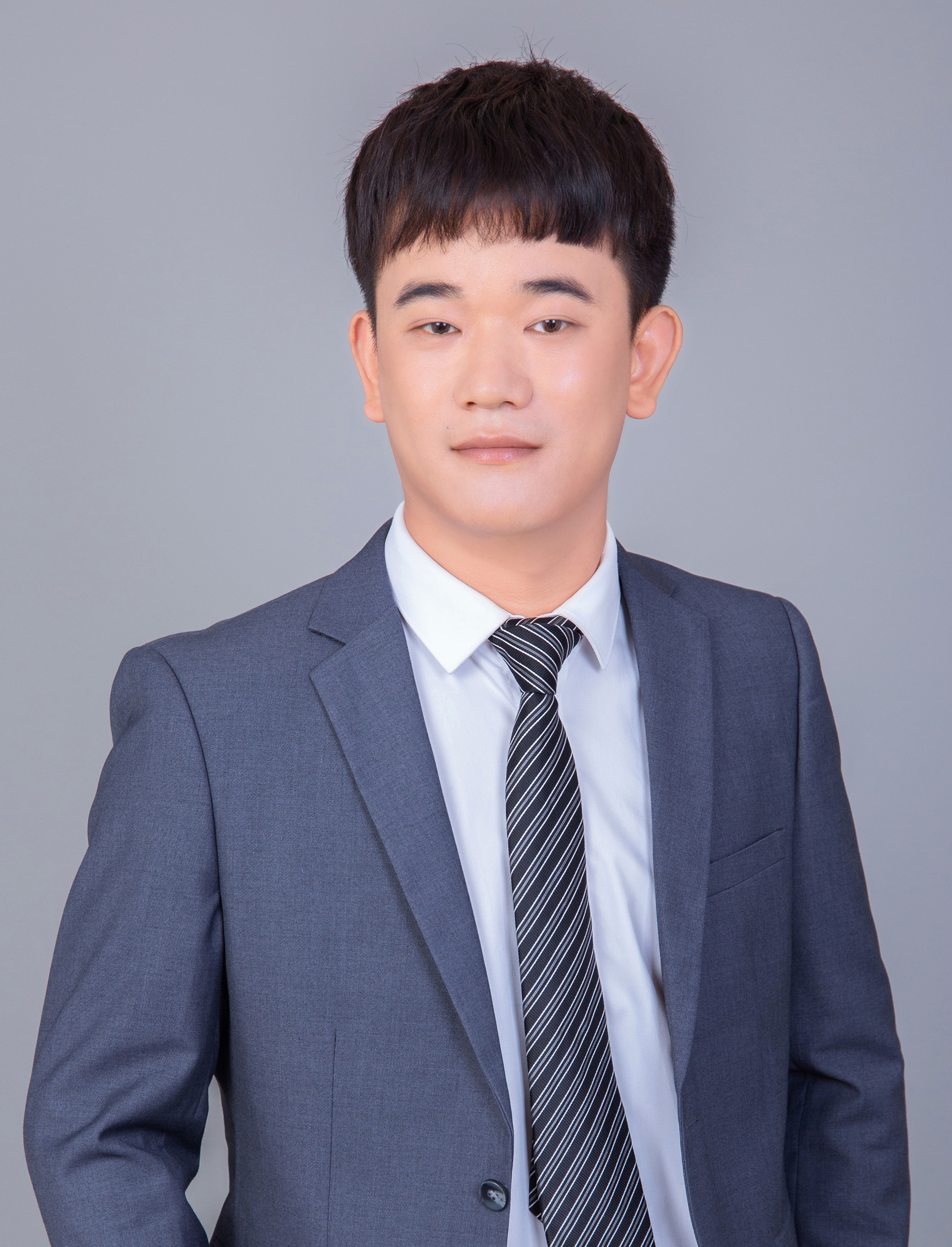}}]
 {Xilong Hou} received the B.Eng. and M.Eng. degrees in mechanical engineering from Northeastern University, Shenyang, China and Harbin Institute of technology (Shenzhen), Shenzhen, China in 2012 and 2015 respectively. Before joining Centre for Artificial Intelligence and Robotics (CAIR) Hong Kong Institute of Science \& Innovation, Chinese Academy of Sciences as an senior engineer, he successively acted as an engineer in ROBO medical, Peng Cheng laboratory and Shenzhen Institute of Artificial Intelligence and Robotics for Society. His research interests include medical robotics and robot assisted surgery.
\end{IEEEbiography}
\vspace{-1.2cm}
\begin{IEEEbiography}[{\includegraphics[width=1in,height=1.25in, clip, keepaspectratio]{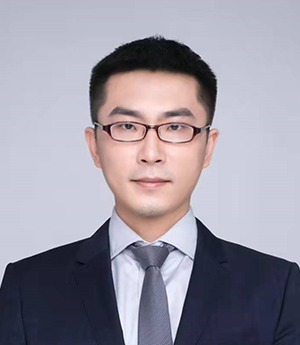}}]
	{Jian Hu} is an Assistant Professor at Chinese Academy of Sciences, Institute of Automation (CASSIA) and the Center of AI and Robotics (CAIR), Hong Kong Institute of Science \& Innovation, Chinese Academy of Sciences. Dr. Hu's research focuses on the development of medical robotic systems with advanced haptic perception.
\end{IEEEbiography}
\vspace{-16.2cm}
\begin{IEEEbiography}[{\includegraphics[width=1in,height=1.45in,keepaspectratio]{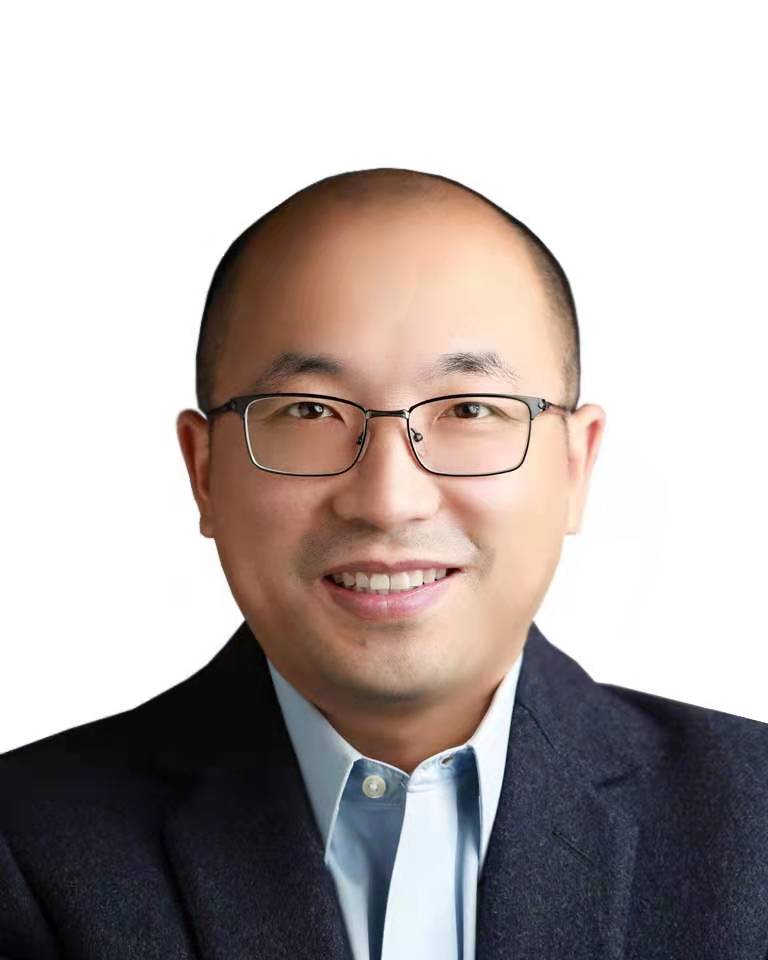}}]{Kang Li} received the Ph.D. degree in Mechanical Engineering from University of Illinois at Urbana Champaign, Champaign, IL, USA, in 2009.  He is now a full professor of the Biomedical Big Data Institute at West China Hospital.  Before joining West China Hospital, He was an associate professor with the Department of Orthopaedics, New Jersey Medical School (NJMS), Rutgers University, Newark, NJ, USA, and an assistant professor with Department of Industrial and Systems Engineering, Rutgers University. His research interests include AI in healthcare, musculoskeletal biomechanics, medical imaging, design and bio-robotics, human reliability, and human factors/ergonomics.
\end{IEEEbiography}

\end{document}